\documentclass[letterpaper]{article} % DO NOT CHANGE THIS
\usepackage{aaai2026}  % DO NOT CHANGE THIS
\usepackage{times}  % DO NOT CHANGE THIS
\usepackage{helvet}  % DO NOT CHANGE THIS
\usepackage{courier}  % DO NOT CHANGE THIS
\usepackage[hyphens]{url}  % DO NOT CHANGE THIS
\usepackage{graphicx} % DO NOT CHANGE THIS
\usepackage{natbib}  % DO NOT CHANGE THIS AND DO NOT ADD ANY OPTIONS TO IT
\usepackage{caption} % DO NOT CHANGE THIS AND DO NOT ADD ANY OPTIONS TO IT
\usepackage{algorithm}
\usepackage{algorithmic}
\usepackage{booktabs}
\usepackage{multirow} 
\usepackage{makecell}
\usepackage{colortbl} 
\usepackage{tikz}
\usepackage{amsmath}
\usepackage{amssymb}
\usepackage{subcaption}
\usepackage{hhline}
\usepackage{enumitem}
\usepackage{pifont}

\usepackage{newfloat}
\usepackage{listings}
\DeclareCaptionStyle{ruled}{labelfont=normalfont,labelsep=colon,strut=off} % DO NOT CHANGE THIS
\floatstyle{ruled}
\newfloat{listing}{tb}{lst}{}
\floatname{listing}{Listing}
\title{Improving Region Representation Learning from Urban Imagery with Noisy Long-Caption Supervision}
\author{
    Yimei Zhang\textsuperscript{\rm 1,2}, 
    Guojiang Shen\textsuperscript{\rm 1,2}, 
    Kaili Ning\textsuperscript{\rm 1,2}, 
    Tongwei Ren\textsuperscript{\rm 3}, \\
    Xuebo Qiu\textsuperscript{\rm 1}, 
    Mengmeng Wang\textsuperscript{\rm 1,2}, 
    Xiangjie Kong\textsuperscript{\rm 1,2,}\thanks{Corresponding author.}
}    
    
\affiliations{
    \textsuperscript{\rm 1} College of Computer Science and Technology, Zhejiang University of Technology\\
     \textsuperscript{\rm 2} Zhejiang Key Laboratory of Visual Information Intelligent Processing\\
    \textsuperscript{\rm 3} State Key Laboratory for Novel Software Technology, Nanjing University\\
    \{yimeizhang0229, gjshen1975, ningkaili,  xueboqiu, wangmengmeng\}@zjut.edu.cn; \\
    rentw@nju.edu.cn; xjkong@ieee.org
    
}

\usepackage{bibentry}
\begin{document}

\maketitle

\begin{abstract}
	Region representation learning plays a pivotal role in urban computing by extracting meaningful features from unlabeled urban data. Analogous to how perceived facial age reflects an individual's health, the visual appearance of a city serves as its ``portrait", encapsulating latent socio-economic and environmental characteristics. Recent studies have explored leveraging Large Language Models (LLMs) to incorporate textual knowledge into imagery-based urban region representation learning. However, two major challenges remain: i)~difficulty in aligning fine-grained visual features with long captions, and ii) suboptimal knowledge incorporation due to noise in LLM-generated captions. To address these issues, we propose a novel pre-training framework called UrbanLN that improves Urban region representation learning through Long-text awareness and Noise suppression. Specifically, we introduce an information-preserved stretching interpolation strategy that aligns long captions with fine-grained visual semantics in complex urban scenes. To effectively mine knowledge from LLM-generated captions and filter out noise, we propose a dual-level optimization strategy. At the data level, a multi-model collaboration pipeline automatically generates diverse and reliable captions without human intervention. At the model level, we employ a momentum-based self-distillation mechanism to generate stable pseudo-targets, facilitating robust cross-modal learning under noisy conditions. Extensive experiments across four real-world cities and various downstream tasks demonstrate the superior performance of our UrbanLN.
\end{abstract}

% Uncomment the following to link to your code, datasets, an extended version or similar.
% You must keep this block between (not within) the abstract and the main body of the paper.
%\begin{links}
%    \link{Code}{https://aaai.org/example/code}
%    \link{Datasets}{https://aaai.org/example/datasets}
%    \link{Extended version}{https://aaai.org/example/extended-version}
%\end{links}

\section{Introduction}
Cities, as vital carriers for living, working, and entertainment, are becoming increasingly complex and diverse due to accelerated urbanization \cite{feng2024citybench}. Consequently, the integration of urban data and advanced deep learning methods for accurate inference of regional attributes has become a pivotal research focus in urban computing, empowering decision-makers with essential insights to guide strategic actions in urban planning, sustainable development, and policy-making \cite{zhou2025urbench, liang2025foundation,zou2025deep}.

Several research lines have emerged to integrate diverse spatial data for urban region representation. POIs have been extensively utilized as proxies for urban functions to assess city vitality, while dynamic population remains a fundamental variable in urban growth theories \cite{zhang2022region, wu2022multi}. However, recent research \cite{fan2023urban} suggests that the visual appearance of the built environment may encapsulate richer information than functional attributes, residential density, or visitor activity patterns. As a result, researchers have increasingly focused on leveraging urban imagery for region representation learning \cite{xi2022beyond,  chen2024profiling, yong2024musecl, hao2025urbanvlp}.

\begin{figure*}[t]
	\centering
	\includegraphics[width=\linewidth]{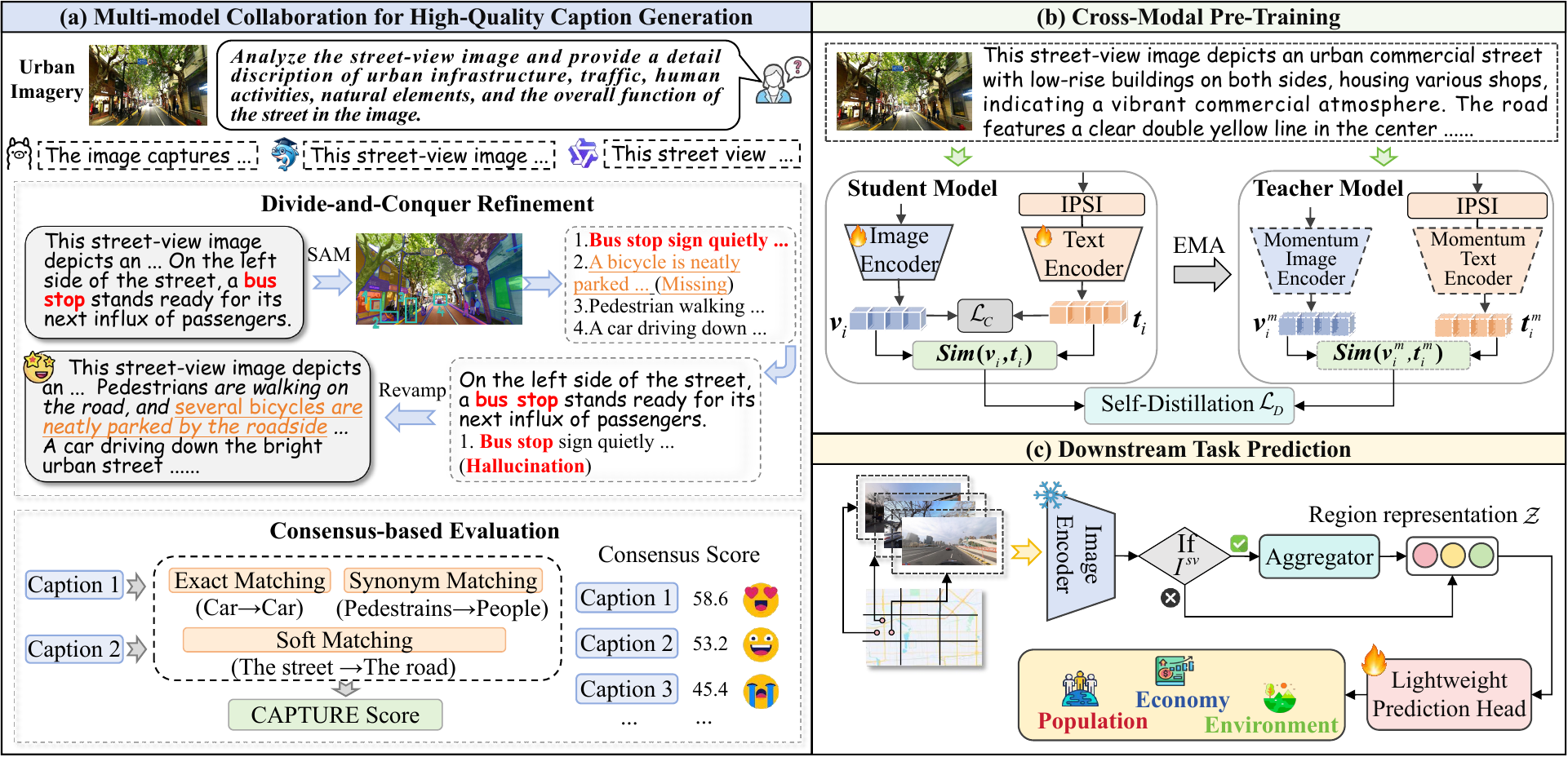}
	\caption{Overview of UrbanLN. The urban imagery input can be either satellite imagery or street-view imagery, with the methodology section primarily focusing on street-view imagery as a representative case.}
	\label{fig1}
\end{figure*}

Previous studies primarily utilize segmentation model \cite{lee2021predicting,fan2023urban} or visual model-based methods for urban imagery feature extraction \cite{huang2023comprehensive}. While effective, these approaches are typically limited in capturing high-level contextual semantics and latent urban knowledge embedded in complex visual scenes. In contrast, the remarkable performance of LLMs across various domains \cite{li2022blip,feng2025urbanllava}, combined with their ability to process and generate language, has positioned them as powerful tools for multimodal learning \cite{li2025causal}. Recent advancements, such as UrbanCLIP \cite{yan2024urbanclip} and UrbanVLP \cite{hao2025urbanvlp}, leverage image captions generated by Multimodal LLMs (MLLMs), enabling deeper insights into urban imagery. Despite these efforts, two key challenges remain: 1)~\textbf{Semantic Bottleneck in Long-Text Processing}: Captions generated by MLLMs embed extensive world knowledge and rich semantic information, providing valuable context for interpreting urban imagery. However, existing methods struggle to effectively process and distill fine-grained features from these often lengthy descriptions. 2)~\textbf{Failure Knowledge Integration from Noisy Captions}: Captions generated by MLLMs often contain noise such as inaccuracies, omissions, or overly generic content \cite{li2025cai}, which hinders effective knowledge integration during cross-modal pre-training. UrbanCLIP addresses this issue via manual calibration, which is time-consuming and not scalable. UrbanVLP introduces scene segmentation ratios to guide caption generation; however, its reliance on fixed category schemas and overemphasis on quantitative cues leads to the loss of high-level semantic information. Furthermore, the use of rigid prompt templates limits linguistic diversity, resulting in homogeneous and low-information descriptions.

In this paper, we propose UrbanLN, a novel cross-modal pre-training framework that enhances urban region representation learning by modeling fine-grained semantics in long-text and mitigating caption noise for more effective knowledge integration. To address \textbf{Challenge 1}, we introduce an Information-Preserved Stretching Interpolation (IPSI) strategy, which enables effective alignment between long-form textual descriptions and fine-grained visual features in intricate urban imagery with minimal additional computational cost.  To address \textbf{Challenge 2}, we propose a dual-level optimization scheme that mitigates noise from LLM-generated captions. At the data level, we integrate multi-MLLM captioning, a divide-and-conquer refinement operation, and consensus-based evaluation, which together promote semantic diversity and improve the reliability of captions. At the model level, we employ a momentum-based self-distillation mechanism, where a momentum-updated version of the student model serves as the teacher to generate pseudo-targets for additional supervision. These pseudo-targets can be seen as diverse ``alternative views'', encouraging the student to learn robust cross-modal representations by differentiating informative cues from noisy content. In summary, this paper makes the following contributions:
\begin{itemize}
	\item We introduce an information-preserved stretching interpolation strategy that relaxes input length constraints with minimal computational overhead. This enables the model to extract fine-grained and holistic features from urban imagery, facilitating more comprehensive urban region representation.
	\item We propose a dual-level optimization strategy to mitigate noise introduced by LLM-generated captions. At the data level, we incorporate multi-MLLM captioning, divide-and-conquer refinement, and consensus-based evaluation to effectively filter out unreliable captions. At the model level, we employ a momentum-driven self-distillation approach to learn semantically robust representations under noisy supervision.
	\item Extensive experiments conducted across four real-world cities and diverse downstream tasks demonstrate that UrbanLN consistently outperforms state-of-the-art methods in urban region representation learning. 
\end{itemize}

\section{Preliminaries}
\textbf{Definition 1 (Urban Region)} \textit{An urban region refers to a spatial unit obtained by partitioning a city based on a specific zoning strategy. Following previous studies \cite{yan2024urbanclip,hao2025urbanvlp}, we obtain the region set $\mathcal{R}$ by partitioning the urban area into a collection of equal-sized, non-overlapping square grids.}

\textbf{Definition 2 (Urban Imagery)}  \textit{Given a city, the urban imagery set is denoted by $\mathcal{I} = \mathcal{I}^{SV} / \mathcal{I}^{SI}$ with the street-view imagery set $\mathcal{I}^{SV}$ and the satellite imagery set $\mathcal{I}^{SI}$.  For each region $R_i \in \mathcal{R}$, it is associated with a set of $u$ street-view images $I_i^{sv} = \{ I_{i,1}^{sv}, I_{i,2}^{sv}, \dots, I_{i,u}^{sv} \} \subset \mathcal{I}^{SV}$ and a satellite image $I_i^{si} \in \mathcal{I}^{SI}$.}

\textbf{Problem Statement (Urban Region Representation Learning)}  \textit{Given the region set $\mathcal{R}$ with its associated urban imagery set $\mathcal{I}$, our goal is to learn the visual representation that serves as region representation, which can be leveraged for a variety of downstream tasks, such as socioeconomic indicator prediction.}

%\textbf{Problem 1 (Urban Region Representation Learning)} \textit{We address this research problem by pre-training image encoders that take urban imagery $\mathcal{I}_i$ associated with a given region $\mathcal{R}_i$ as input and produce high-dimensional embeddings. These embeddings serve as compact and transferable region representations that support various downstream tasks, such as socioeconomic indicator prediction.}

\section{Methodology}
As illustrated in Figure \ref{fig1}, UrbanLN comprises three key components: 1) a multi-model collaboration caption generation pipeline to produce high-quality texts; 2) a cross-modal pre-training framework designed for efficient training of a powerful urban image encoder; and 3) a prediction module for various downstream tasks by simply fine-tuning a lightweight prediction head.

\subsection{Multi-model Collaboration for High-Quality Caption Generation}
To achieve a more comprehensive understanding of urban imagery, we first leverage MLLMs to generate semantically rich and context-aware descriptions for each image. These captions go beyond surface-level visual elements by incorporating the model's world knowledge and reasoning capabilities, capturing key attributes that reflect the characteristics of urban environments. However, captions generated by MLLMs are often prone to hallucinations and information homogenization \cite{ye2025claim}. To tackle this issue, we propose a multi-model collaboration approach aimed at generating diverse, comprehensive, and reliable captions by employing three key strategies as follows.

\subsubsection{Multi-MLLM Captioning.} We utilize multiple MLLMs, where each model independently serves as a captioning agent to generate multiple long captions. The diversity in different MLLMs not only mitigates potential semantic shifts introduced by individual models but also generates rich textual data across the entire dataset, which can be seen as a form of text data augmentation, similar to language rewriting \cite{fan2023improving}, thereby promoting subsequent cross-modal pre-training.

\subsubsection{Divide-and-Conquer Refinement.} 
To alleviate the issues of hallucinations and visual element omissions, we incorporate fine-grained visual details to reinforce the completeness and accuracy of the captions. Firstly, we employ the Segment Anything Model (SAM) \cite{kirillov2023segment} for image segmentation, and apply a maximal rectangle algorithm to reduce overlap among the remaining regions, resulting in a set of cropped bounding boxes that represent salient visual elements. Secondly, for each detected salient visual element, we utilize the respective MLLM to generate concise local short captions that complement the long caption by addressing potentially omitted visual details. Thirdly, to further suppress hallucinations, we introduce a phrase-level filtering mechanism. Visual element phrases are first extracted from both original long and local short captions using the Factual parser \cite{li2023factual}, which is a state-of-the-art model for text scene graph parsing. These phrases are then scored by OWLv2 \cite{minderer2023scaling}, an open-vocabulary object detection model. Phrases with a score below 0.01 are discarded, reducing the inclusion of hallucinated content. Finally, we reuse the same MLLM that reasonably absorbs detailed information from local short captions based on image content and original long caption structures to generate more complete and natural image descriptions.

\subsubsection{Consensus-based Evaluation.} 
Following the above refinement process, the quality of the caption is improved. However, in the absence of ground-truth captions, existing quantitative metrics fall short in accurately assessing their quality. Inspired by real-world decision-making processes, where consensus among multiple capable and independent individuals is often considered reliable, we propose a consensus-based evaluation method. This method uses inter-model consistency as an indicator of caption reliability under unsupervised settings. 

Specifically, we adopt the CAPTURE metric \cite{dong2024benchmarking} to measure the similarity between two candidate captions, as it leverages structured visual semantics extracted by the Factual parser, enabling fine-grained alignment of objects, attributes, and relations, rather than relying on generic sentence-level embeddings. Following its procedure, we extract core information such as objects, attributes, and relations from the descriptions using the Factual parser, and remove redundant or irrelevant content. Then, for the extracted core information, using three matching strategies: 1) exact matching (checking if the core elements in the two captions match); 2) synonym matching (checking if two candidate captions' synonym sets overlap);  and 3) soft matching (encoding and computing the similarity for remaining unmatched phrases). Next, calculating the precision and recall rates for each type of core information, where the soft matching score is used as a complement to exact matched and synonym matched relationships. More details can be found in \cite{dong2024benchmarking}. Finally, the CAPTURE metric is obtained as follows:
\begin{equation}
	\text{CAPTURE} = \frac{\alpha F1_{obj} + \beta F1_{attr} + \gamma F1_{rel}}{\alpha + \beta + \gamma},
	\label{eq1}
\end{equation}
where $\alpha$, $\beta$ and $\gamma$ are scale factors, and $F1_*$ stands for the $F1$ score of  each type of core information.

The consensus score of each caption is then computed by averaging its CAPTURE scores with all other captions generated for the same image. A higher consensus score indicates stronger alignment with the shared understanding across multiple MLLMs, reflecting greater inter-model agreement. Finally, we select the caption candidate with the highest consistency score for each image to obtain a high-quality urban image-text pair dataset, denoted as $\mathcal{D} = \{\mathcal{I}, \mathcal{T}\}$.

\subsection{Cross-Modal Pre-Training}
To effectively transfer the knowledge embedded in the long captions into the visual representations, we adopt the Contrastive Language-Image Pre-training (CLIP) model \cite{radford2021learning} as the backbone to align text and image.

\subsubsection{Information-Preserved Stretching Interpolation.} Although CLIP has demonstrated strong capabilities in cross-modal representation learning, the token limitation of its text encoder (77 tokens) drastically reduces its performance in processing long text inputs \cite{wu2024lotlip, zheng2024dreamlip,zhang2024long}. This limitation severely constrains its application in tasks requiring complex textual descriptions, especially in fine-grained urban region representation learning, where the average length of each caption exceeds 100 words. Consequently, directly applying CLIP to our task is not impractical.

To overcome CLIP's limitations in handling long texts, we propose an Information-Preserved Stretching Interpolation (IPSI) strategy. This approach is inspired by prior studies \cite{zhang2024long}, which demonstrate that the lower positional embeddings (i.e., the first 20) in CLIP are well-trained and effectively capture absolute positional information. Directly interpolating these embeddings may disrupt the established positional representation, resulting in degraded performance and higher computational costs. To balance stability and computational efficiency, we preserve the first 20 positions and apply interpolation only to the remaining 57 positions. The detailed process is as follows:
\begin{equation}
	P^*(e) = 
	\begin{cases} 
		P(e), & e \leq 20; \\ 
		(1-\omega) P\left(\left\lfloor \frac{e}{\lambda} \right\rfloor\right) + \omega P\left(\left\lceil \frac{e}{\lambda} \right\rceil\right),  & \text{otherwise},
	\end{cases}
	\label{eq2}
\end{equation}
where $\lambda$ is the interpolation ratio, and $\omega = \frac{e}{\lambda} - \left\lfloor \frac{e}{\lambda} \right\rfloor$, which is used as a weighting factor to ensure smooth transition of the new position.

\subsubsection{Momentum-based Self-Distillation.}
Despite our efforts to improve caption quality, errors or hallucinations remain non-trivial issues. Furthermore, due to the high visual similarity among certain urban neighborhoods, some captions selected as negatives may still accurately describe the visual content of the corresponding urban images. To tackle these issues, we introduce a momentum-based self-distillation method designed to improve learning under noisy supervision. This method keeps a momentum version \cite{he2020momentum, li2021align} of the student model as the teacher. During training, the teacher's parameters are updated at each iteration using the Exponential Moving Average (EMA) of the student's, and the student minimizes a contrastive loss $\mathcal{L}_C$, while aligning its outputs with those of the teacher via a distillation loss $\mathcal{L}_D$.

Specifically, we maintain two dynamic queues $\{\boldsymbol{v}^m_k, \boldsymbol{t}^m_k\}_{k \in K}$ that store the most recent $K$ image-text representations encoded by the teacher model. Then, the similarity of the image-text and text-image can be formulated as:
\begin{equation}
	\begin{split}
		&p^{\text{i2t}}_k(I_i) = \frac{\exp(s_{\boldsymbol{v}_i, \boldsymbol{t}_k} / \tau)}{\sum_{k=1}^{K+N} \exp(s_{\boldsymbol{v}_i, \boldsymbol{t}_k} / \tau)}, \\
		&p^{\text{t2i}}_k(T_i) = \frac{\exp(s_{\boldsymbol{t}_i, \boldsymbol{v}_k} / \tau)}{\sum_{k=1}^{K+N} \exp(s_{\boldsymbol{t}_i, \boldsymbol{v}_k} / \tau)},
		\label{eq3}
	\end{split}
\end{equation}
where $N$ is the batch size, $s_{\boldsymbol{v}_i,\boldsymbol{t}_k} =\cos\langle \boldsymbol{v}_i, \boldsymbol{t}_k^m\rangle$, $s_{\boldsymbol{t}_i, \boldsymbol{v}_k} =\cos\langle \boldsymbol{t}_i$,$\boldsymbol{v}_k^m \rangle$, and $\tau$ is a learnable temperature parameter. The contrastive loss is:
\begin{equation}
	\mathcal{L}_C = -\frac{1}{2N} (\sum_{i=1}^N \log p^{\text{i2t}}(I_i) + \sum_{i=1}^N \log p^{\text{t2i}}(T_i)).
	\label{eq5}
\end{equation}
Next, we incorporate a distillation mechanism that allows the student to learn from the teacher’s more stable pseudo-targets. We firstly compute the image-text similarity  $s_{\boldsymbol{v}_i, \boldsymbol{t}_i}^m = \cos\langle \boldsymbol{v}_i^m, \boldsymbol{t}_i^m \rangle$ from the teacher, and obtain pseudo-targets $q^\text{i2t}$ and $q^\text{t2i}$ using the same formulation as Eq. (\ref{eq3}) but replacing $s$ in the original formulation with $s^m$. The distillation loss is then computed as:
\begin{equation}
	\begin{split}
		\mathcal{L}_D = \frac{1}{2} \mathbb{E}_{(I, T) \sim \mathcal{D}} 
		[ 
		& \text{KL}\left(q^{\text{i2t}}(I) \| p^{\text{i2t}}(I)\right) \\
		+ & \text{KL}\left(q^{\text{t2i}}(T) \| p^{\text{t2i}}(T)\right) 
		],
	\end{split}
	\label{eq6}
\end{equation}
where KL denotes the Kullback-Leibler divergence. The pseudo-targets generated by the momentum teacher offer additional perspectives beyond the original image-text pairs, guiding the student to learn representations that are robust to semantic-preserving transformations under noisy conditions. This encourages the model to capture shared semantic clues across diverse samples rather than overfitting noise.  The final loss for pre-training is: 
\begin{equation}
	\mathcal{L} = (1 - \mu)\mathcal{L}_C +\mu \mathcal{L}_D ,	
	\label{eq7}
\end{equation}
where $\mu \in(0,1)$ controls the balance between contrastive learning and teacher-guided supervision.

\subsection{Downstream Task Prediction}
Through optimizing the loss function in Eq.~(\ref{eq7}), we obtain a knowledge-enhanced image encoder. Given any urban imagery $I^{sv}_i$ or $I^{si}_i$ corresponding to a specific region $R_i$, we use the frozen image encoder to extract a compact feature representation $\boldsymbol{z}_i$, which serves as the urban region representation\footnote{For multiple street-view images of the same region, we use average pooling of visual features to represent the region.}. For each downstream task, we fine-tune a lightweight prediction head (e.g., a multi-layer perceptron) to produce the output as $\hat{Y}_i = MLP(\boldsymbol{z}_i)$.

\begin{table*}[t]
	\small
	\setlength{\tabcolsep}{2pt}
	\centering
	\begin{tabular}{l|ccc|ccc|ccc|ccc|ccc}
		\toprule 
		%		\rowcolor{gray!20}
		\multirow{1}{*}{\textbf{Task $\rightarrow$}}  & \multicolumn{3}{c}{\textbf{Pop}}   & \multicolumn{3}{c}{\textbf{GDP}} & \multicolumn{3}{c}{\textbf{Night}} & \multicolumn{3}{c}{\textbf{Com}}  & \multicolumn{3}{c}{\textbf{CO$_2$}} \\ 	\cmidrule(lr){2-4} \cmidrule(lr){5-7} \cmidrule(lr){8-10} \cmidrule(lr){11-13} \cmidrule(lr){14-16} 
		%		\rowcolor{gray!20}
		\multirow{1}{*}{\textbf{Model $\downarrow$}}           & $R^2$$\uparrow$ & RMSE$\downarrow$& MAE$\downarrow$ & $R^2$$\uparrow$& RMSE$\downarrow$& MAE$\downarrow$ & $R^2$$\uparrow$ & RMSE$\downarrow$ & MAE$\downarrow$ & $R^2$$\uparrow$ & RMSE$\downarrow$ & MAE$\downarrow$ & $R^2$$\uparrow$ & RMSE$\downarrow$ & MAE$\downarrow$         \\ \midrule
		ResNet-18& 0.323& 0.912 & 0.728  &   0.093 & 1.702 & 1.402  & 0.105 & 0.875 & 0.744  &   0.255 & 2.612 & 2.082        &     0.312 & 0.725 & 0.521  \\
		Tile2Vec & 0.446  & 0.903    & 0.716   &  0.165   & 1.565 & 1.350  & 0.268 & 0.660    & 0.547  &  0.296   & 2.489    &2.102  &  0.336   & 0.665  & 0.517  \\
		PG-SimCLR & 0.549  & 0.802  &  0.612  & 0.316  & 1.450  & 1.163  & 0.334  &  0.669 &  0.538  &    0.429      &   2.337      &   1.862      &     0.467     &   0.642      &    \textit  0.452         \\
		RemoteCLIP  & 0.422  & 0.908  &   0.684  & 0.245  &  1.524 & 1.241  & 0.128  & 0.766  & 0.607  &     0.338     &    2.516     &  2.059        &   0.378       &  0.693       & 0.501           \\
		UrbanCLIP&  0.595 & 0.761   & 0.556    & 0.358   & 1.410 & 1.121 &  0.426    &0.621   & 0.503   &    0.504   & 2.226   & 1.764      &  0.480 & 0.619   & 0.492       \\ 
		UrbanVLP &  \textit {0.619}  &  \textit{0.728} & \textit{0.524}   & \textit{0.372}  & \textit{1.390 }   & \textit{1.108}   &  \textit{  0.454}  &  \textit{ 0.606}   & \textit{  0.493}   &  \textit{ 0.555}    & \textit{ 2.108}    & \textit{ 1.687}      &  \textit{  0.487}   & \textit  {0.615}  &   0.492           \\ 
		\midrule
		\textbf{UrbanLN+SV}  &  \textbf {0.705} &	  \textbf{ 0.686} &	  \textbf {0.491} &	  \textbf {0.440}&	  \textbf{ 1.373}&	  \textbf{ 1.043} &	\underline{0.514} &\underline{0.562} &	\underline{0.449}&\underline{0.591} &	  \textbf{1.976} &\underline{	1.572} &\underline{	0.677} &	\underline{0.496}&	\underline{0.390} \\
		Improvement &   13.9\%&5.8\%& 6.3\% & 18.3\%&1.2\%& 5.9\%&13.2\% & 7.3\%&8.9\% & 6.5\%& 6.3\%& 6.8\%&39.0\%&19.3\% & 13.7\%   \\
		
		\textbf{UrbanLN+SI}   &\underline{   0.686} &\underline{0.714} &\underline{	  0.496} &\underline{	  0.437} &	\underline {  1.380} &\underline{	  1.057} &  \textbf	{0.519} &  \textbf{	0.553}&  \textbf{0.447} &  \textbf{	0.597} &\underline{ 2.005}&	  \textbf{1.568} &  \textbf{	0.688} &	  \textbf{0.478} &  \textbf{0.383 }\\
		
		Improvement &   10.8\%	& 1.9\%	&5.3\%	& 17.5\%& 	0.7\%	& 4.6\%	& 14.3\%	& 8.7\%	& 9.3\%	& 7.6\%	& 4.9\%	& 7.1\%	& 41.3\%	& 22.3\%	& 15.3\%
		\\ 
		\bottomrule
	\end{tabular}
	\caption{Downstream tasks prediction results on the BJ dataset. The abbreviations ``+SV" and ``+SI" denote the inclusion of street-view images and satellite images as urban imagery. The best, second-best, and third-best results are highlighted in bold, underlined, and italic, respectively. The ``Improvement" row indicates the relative gain (\%) over the best-performing baseline.}
	\label{table2}
\end{table*}

\section{Experiment}
\subsection{Experimental Setup}

\subsubsection{Datasets.} 
We collect satellite and street-view images for Beijing (BJ), Shanghai (SH), and Shenzhen (SZ) from Baidu Maps. The corresponding data for New York (NY) are sourced from \cite{yong2024musecl}. For downstream tasks, we evaluate our UrbanLN on the following urban indicators: population (\textbf{Pop}), gross domestic product (\textbf{GDP}), nighttime light intensity (\textbf{Night}), number of restaurant comments (\textbf{Com}), carbon emissions ($\mathbf{CO_2}$), number of POIs (\textbf{POI}), and crime incidence (\textbf{Crime}). More details can be found in the Appendix. All indicator data have been transformed into the logarithmic scale using the formula $Y = \ln(Y_{\text{original}} + 1)$. In the prediction phase, every downstream task dataset is randomly divided into training, validation, and test sets in a ratio of 6:2:2.

\subsubsection{Implementation Details.} We employ several MLLMs for long caption generation, including LLaMA-Adapter V2 \cite{gao2023llama}, ShareGPT4V-7B \cite{chen2024sharegpt4v}, Qwen2.5-VL-7B \cite{wang2024qwen2}, DeepSeek-VL2-tiny \cite{wu2024deepseek}, and InternVL3-8B \cite{zhu2025internvl3}. We set $\alpha,\beta,\gamma=5,5,2$ following the default settings in \cite{dong2024benchmarking}. During pre-training, we use the CLIP model with ViT-B/16 backbone. The interpolation ratio $\lambda$ is set to 4, extending the maximum input length to 248, and the weight $\mu$ is set to 0.5. The momentum parameter for updating the momentum teacher and the queue length are empirically set to 0.995 and 4096.  To optimize UrbanLN, we utilize the AdamW optimizer and initialize the learning rate at $1e\text{-}7$.

\begin{table}[t]
	\centering
	\small
	\begin{tabular}{l|cccc}
		\toprule
		\textbf{Model} & \textbf{Pop} & \textbf{GDP} & \textbf{Night}  & \textbf{CO$_2$} \\
		\midrule
		ResNet-18 & 0.454 & 0.281 & 0.275 & 0.454 \\
		Tile2Vec & 0.482& 0.351 & 0.316 & 0.542 \\
		PG-SimCLR & 0.544  & 0.472 & 0.348 & 0.568 \\
		RemoteCLIP & 0.524  & 0.395 & 0.391 & 0.564 \\
		UrbanCLIP & 0.554 & \textit{0.473} & 0.464 & 0.615 \\
		UrbanVLP &  \textit{0.576}  & 0.458 & \textit{0.506} & \textit{0.637} \\
		\midrule
		\textbf{UrbanLN+SV} & \textbf{0.669} & \textbf{0.480} & \underline{0.552}   &  \underline{0.697}  \\
		Improvement &16.1\%		&1.5\%	&	9.1\%	&9.4\% \\
		\textbf{UrbanLN+SI} &  \underline{0.661}  &  \underline{0.476}  & \textbf{0.555} & \textbf{0.718} \\
		Improvement &14.8\%	&0.6\%	&	9.7\%	&	12.7\% \\
		\bottomrule
	\end{tabular}
	\caption{Prediction results on the SH dataset ($R^2$ metric).}
	\label{tab:sh_results}
\end{table}

\begin{table}[t]
	\centering
	\small
	\begin{tabular}{l|ccc}
		\toprule
		\textbf{Model} & \textbf{Pop} &\textbf{Crime}    & \textbf{POI} \\
		\midrule
		ResNet-18 & -0.079    & 0.050   & 0.160\\
		Tile2Vec  &0.096     & 0.135  &0.192 \\
		PG-SimCLR  & 0.375    &0.196   &--\\
		RemoteCLIP & 0.364      &0.154  &0.328 \\
		UrbanCLIP  &0.448      &0.251 &0.425 \\
		MuseCL& 0.521      &\textit{0.368}& \textit{0.542} \\
		UrbanVLP & \textit{0.534}    & \underline{0.467}&  \underline{0.583} \\
		\midrule
		\textbf{UrbanLN+SV} &  \underline{0.676}    &\textbf{0.723}	&\textbf{0.650} \\
		Improvement &26.6\%   &54.8\%&11.5\% \\
		\textbf{UrbanLN+SI} &\textbf{0.725}    &0.323 &	0.521 \\
		Improvement &35.8\%		&-30.8\%&	-10.6\% \\
		\bottomrule
	\end{tabular}
	\caption{Prediction results on the NY dataset ($R^2$ metric).}
	\label{tab:ny_results}
\end{table}

\subsubsection{Baselines.}We compare UrbanLN with a variety of baselines commonly used in imagery-based urban region representation learning. These include ResNet-18 \cite{he2016deep}, Tile2Vec \cite{jean2019tile2vec}, PG-SimCLR \cite{xi2022beyond}, RemoteCLIP \cite{liu2024remoteclip}, UrbanCLIP \cite{yan2024urbanclip}, MuseCL \cite{yong2024musecl} and UrbanVLP \cite{hao2025urbanvlp}. MuseCL is restricted to the NY dataset owing to missing mobility data in BJ, SH, and SZ. The details of the baselines are in the Appendix.

\subsubsection{Evaluation Metrics.} For downstream tasks evaluation, we employ widely used metrics including coefficient of determination ($R^2$), Root Mean Squared Error (RMSE), and Mean Absolute Error (MAE).

\begin{figure}[t]
	\centering
	\begin{subfigure}[t]{0.15\textwidth}  
		\includegraphics[width=\linewidth]{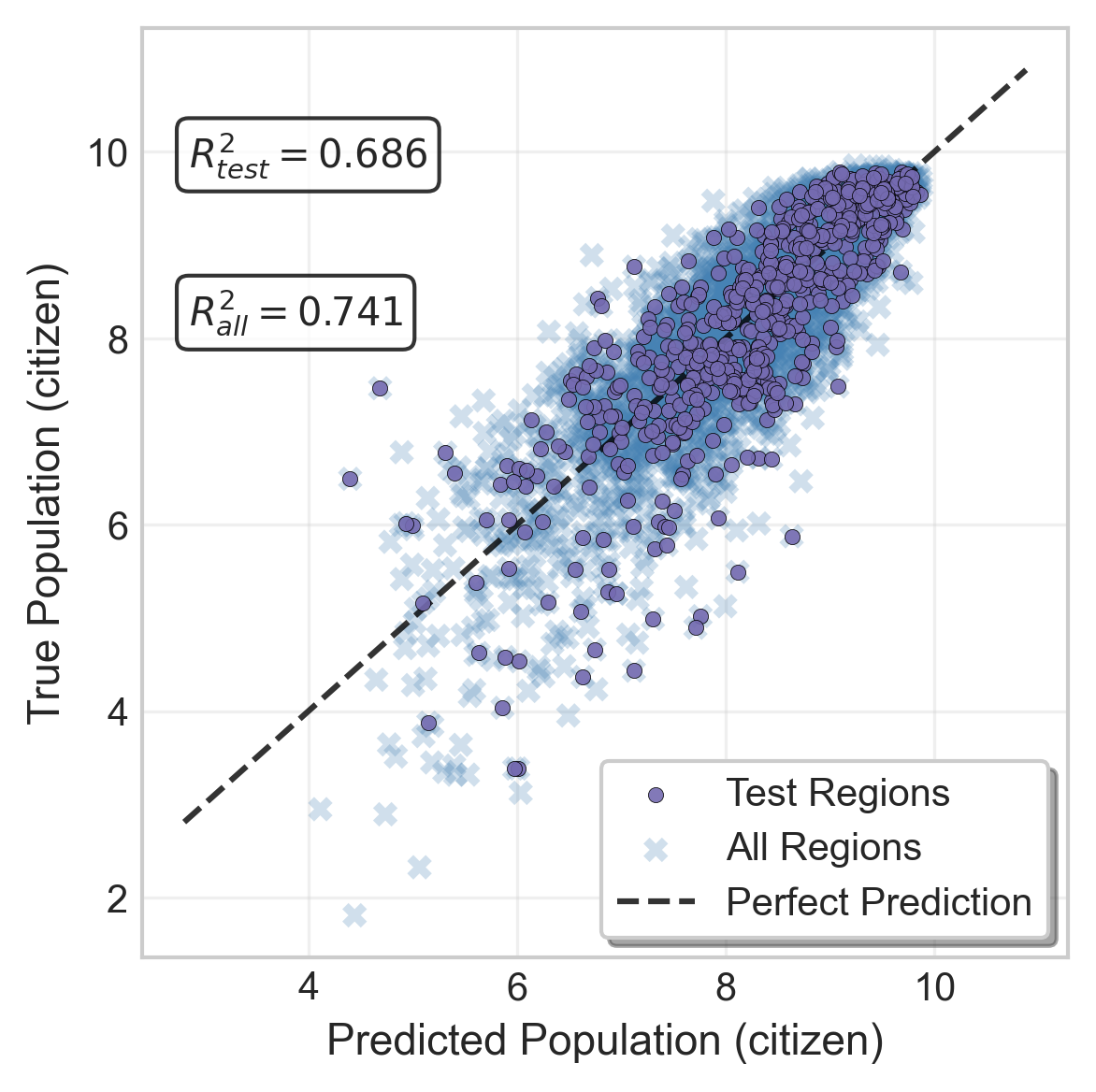}
		\caption{Pop.}
	\end{subfigure}
	\hfill  
	\begin{subfigure}[t]{0.15\textwidth}
		\includegraphics[width=\linewidth]{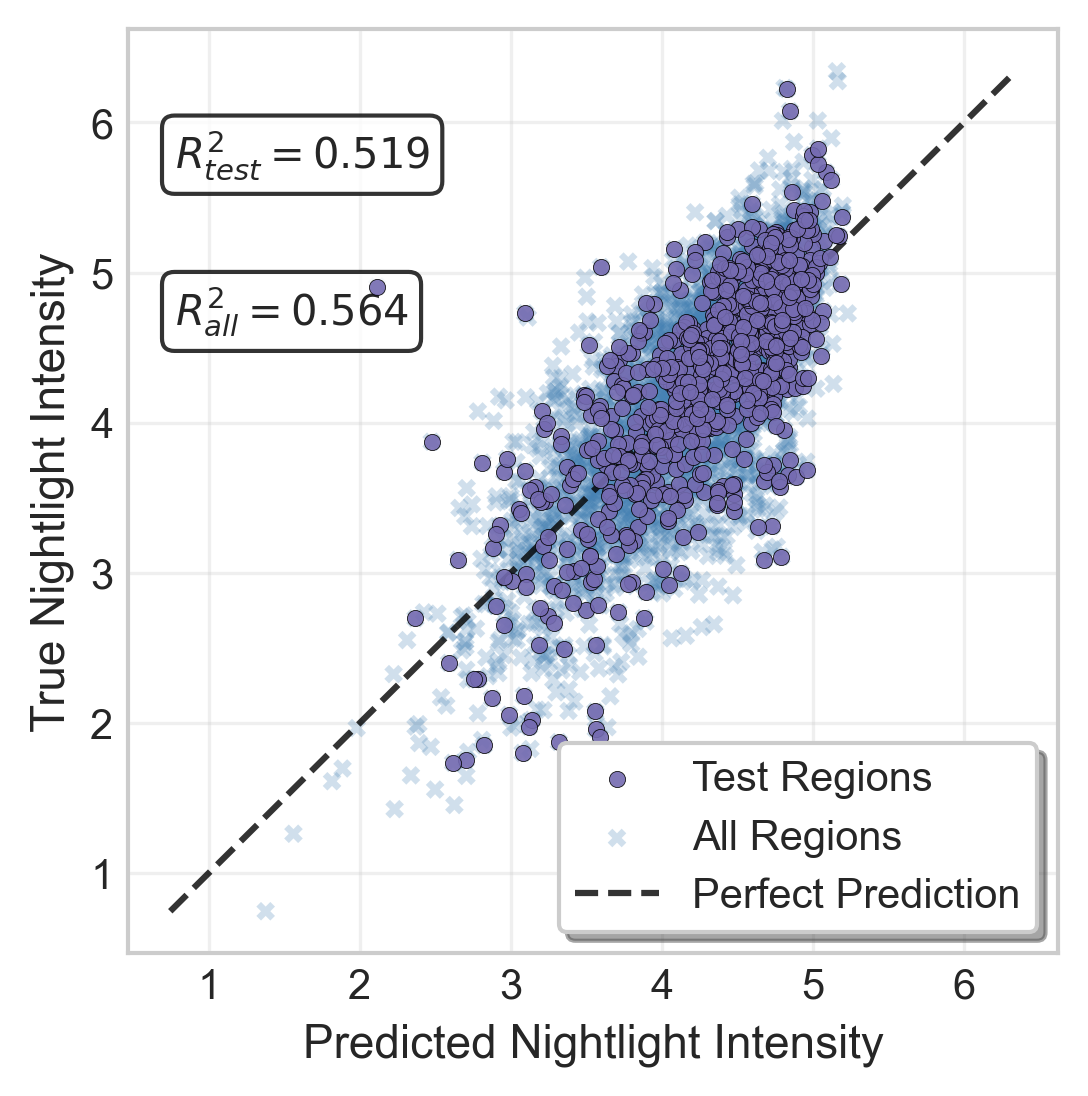}
		\caption{Night.}
	\end{subfigure}
	\hfill  
	\begin{subfigure}[t]{0.15\textwidth}
		\includegraphics[width=\linewidth]{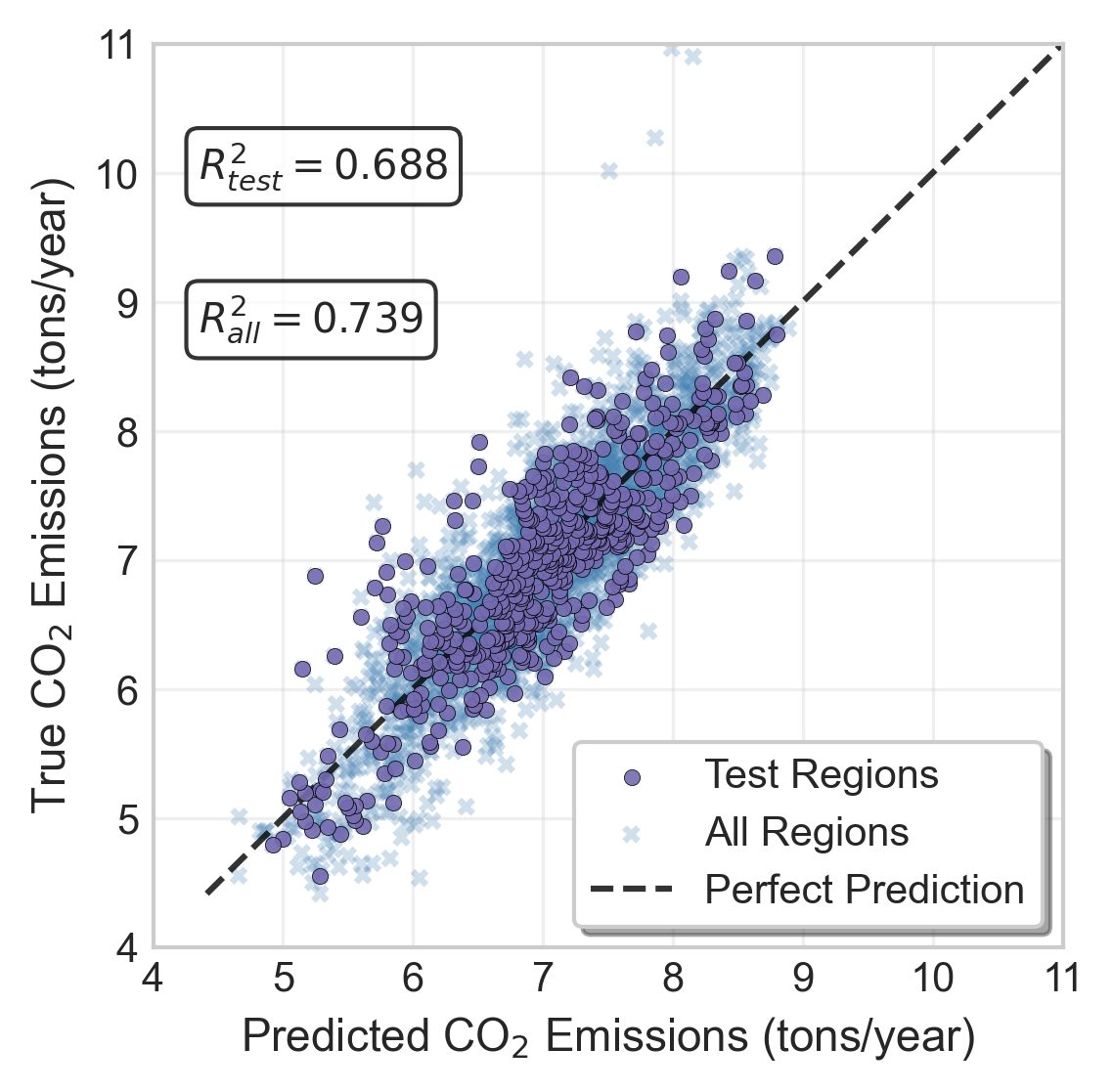}
		\caption{$\text{CO}_2$.}
	\end{subfigure}
	\caption{Prediction versus the ground truth on the BJ dataset using satellite imagery. The dotted line is at 45$^\circ$. $R^2_{test}$ and $R^2_{all}$ correspond to the results of testing regions (purple dots) and all regions (blue crosses), respectively.}
	\label{fig_r2}
\end{figure}

\subsection{Comparative Results}
The experimental results on the BJ dataset are shown in Table~\ref{table2}. Compared with the strongest baselines, the proposed UrbanLN achieves average improvements of 18.23\%, 7.84\%, and 8.32\% in terms of $R^2$, RMSE, and MAE, respectively. PG-SimCLR introduces POI semantic similarity and geographical proximity, thus yielding moderate improvements. RemoteCLIP, a foundation model for remote sensing images, excels in image-text retrieval and object counting but shows limited capability in urban understanding tasks. UrbanCLIP and UrbanVLP enhance image representations through textual description; however, their performance is constrained by limited semantic extraction, as well as hallucinations and homogenized expressions in the generated captions. By contrast, UrbanLN enables the input of long texts to capture fine-grained features in urban imagery, facilitating a more comprehensive understanding of urban environments. Furthermore, the proposed noise optimization strategy is applied at both the data and model levels, which enhances UrbanLN’s ability to integrate knowledge from LLM-generated captions, thereby contributing to its superior performance. To further assess the predictive power of UrbanLN, we compare the predicted and true urban indicators for all regions in the BJ dataset. As shown in Figure \ref{fig_r2}, UrbanLN closely replicates the true values for most regions.

We also include datasets from additional cities to evaluate the adaptability of our model. The $R^2$ results for SH and NY are summarized in Tables \ref{tab:sh_results} and \ref{tab:ny_results}. Due to space limitations, results on additional evaluation metrics and the SZ dataset are provided in the \textit{Appendix}. Our model consistently demonstrates superior performance across all downstream tasks in both SH and SZ. Notably, on the NY dataset, the UrbanLN+SV delivers the best results across all prediction tasks, with an average improvement of 30.97\%. In contrast, UrbanLN+SI shows relatively suboptimal results in crime and POI prediction tasks. This performance gap may stem from the inherent limitations of satellite imagery and the lower spatial granularity of this dataset. By comparison, baselines such as MuseCL and UrbanVLP, which incorporate large-scale street-view images, achieve competitive performance, benefiting from their enhanced ability to model internal urban structures and human activity patterns.

\begin{figure}[t]
	\centering
	\begin{subfigure}[t]{0.49\linewidth}
		\includegraphics[width=\linewidth]{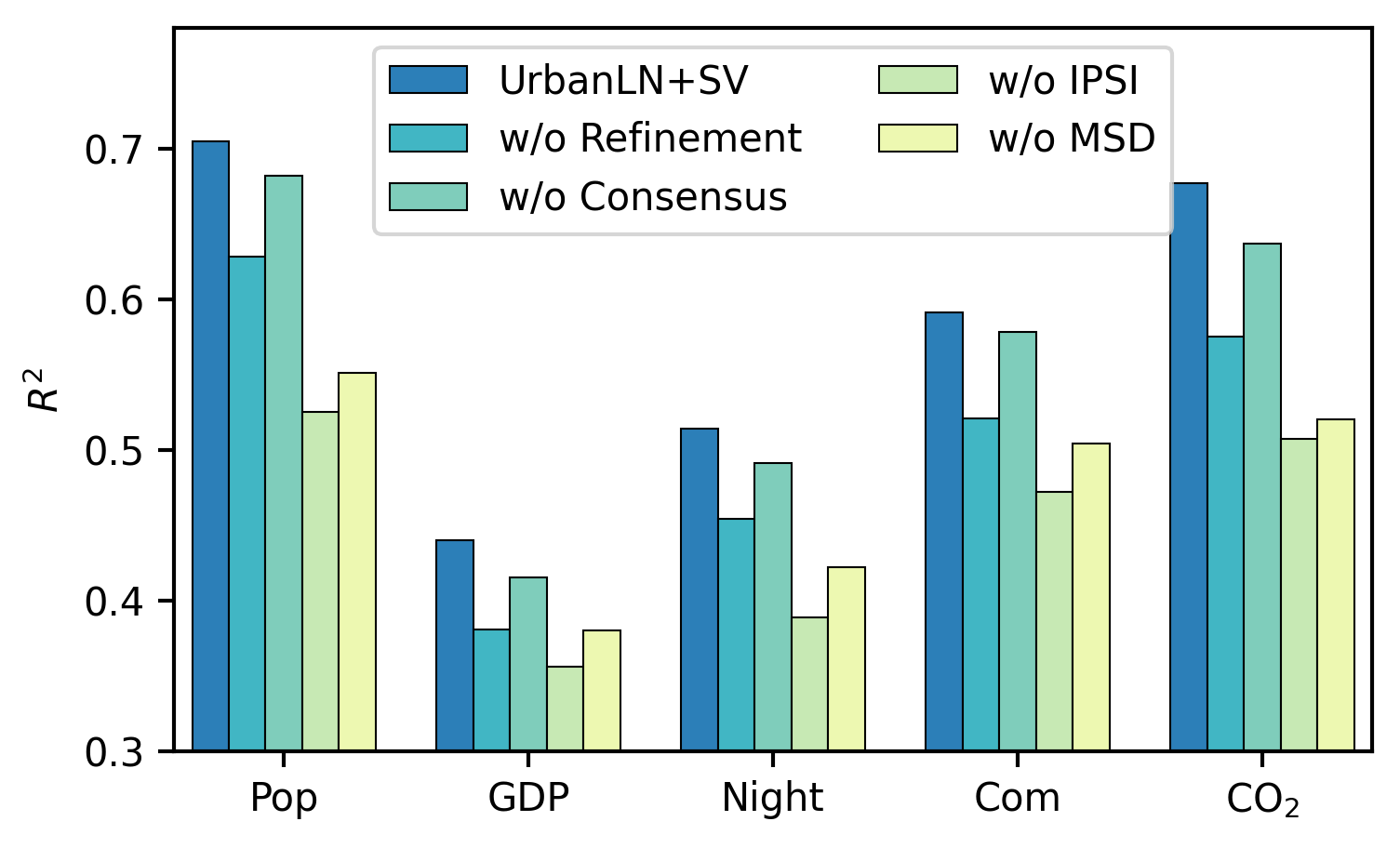}
		\caption{Street-view imagery on BJ.}
	\end{subfigure}
	\begin{subfigure}[t]{0.49\linewidth}
		\includegraphics[width=\linewidth]{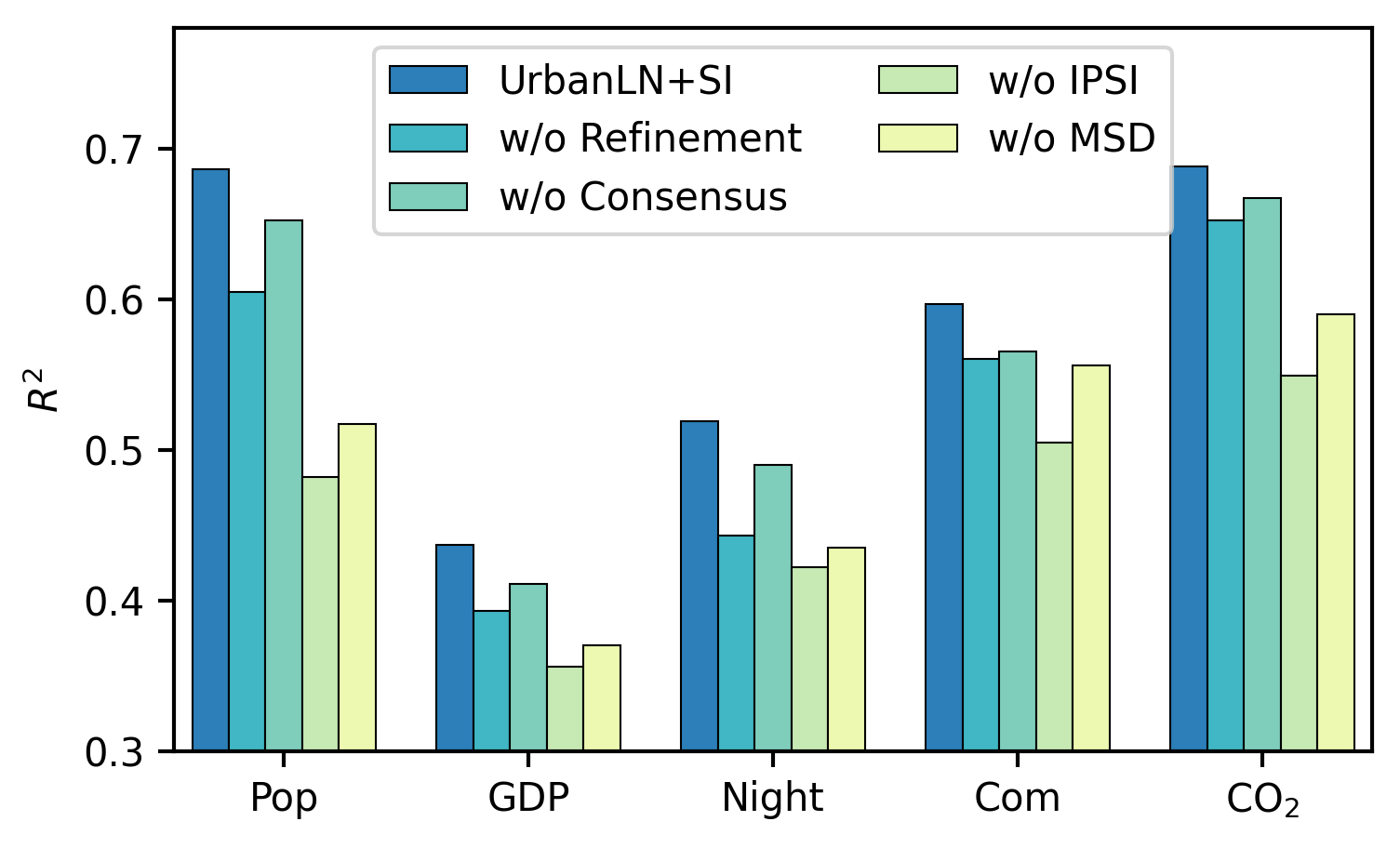}
		\caption{Satellite imagery on BJ.}
	\end{subfigure}
	
	\begin{subfigure}[t]{0.49\linewidth}
		\includegraphics[width=\linewidth]{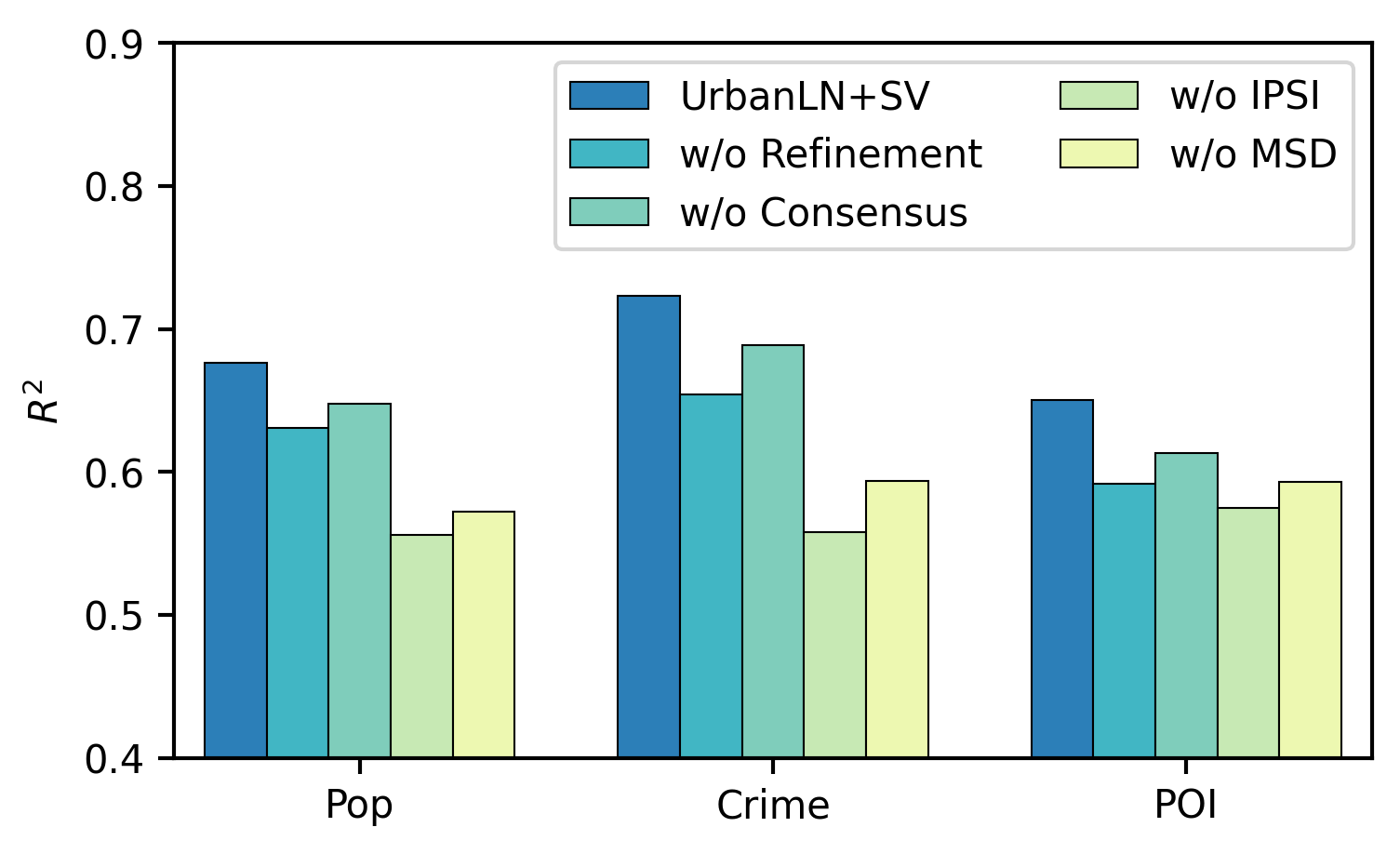}
		\caption{Street-view imagery on NY.}
	\end{subfigure}
	\begin{subfigure}[t]{0.49\linewidth}
		\includegraphics[width=\linewidth]{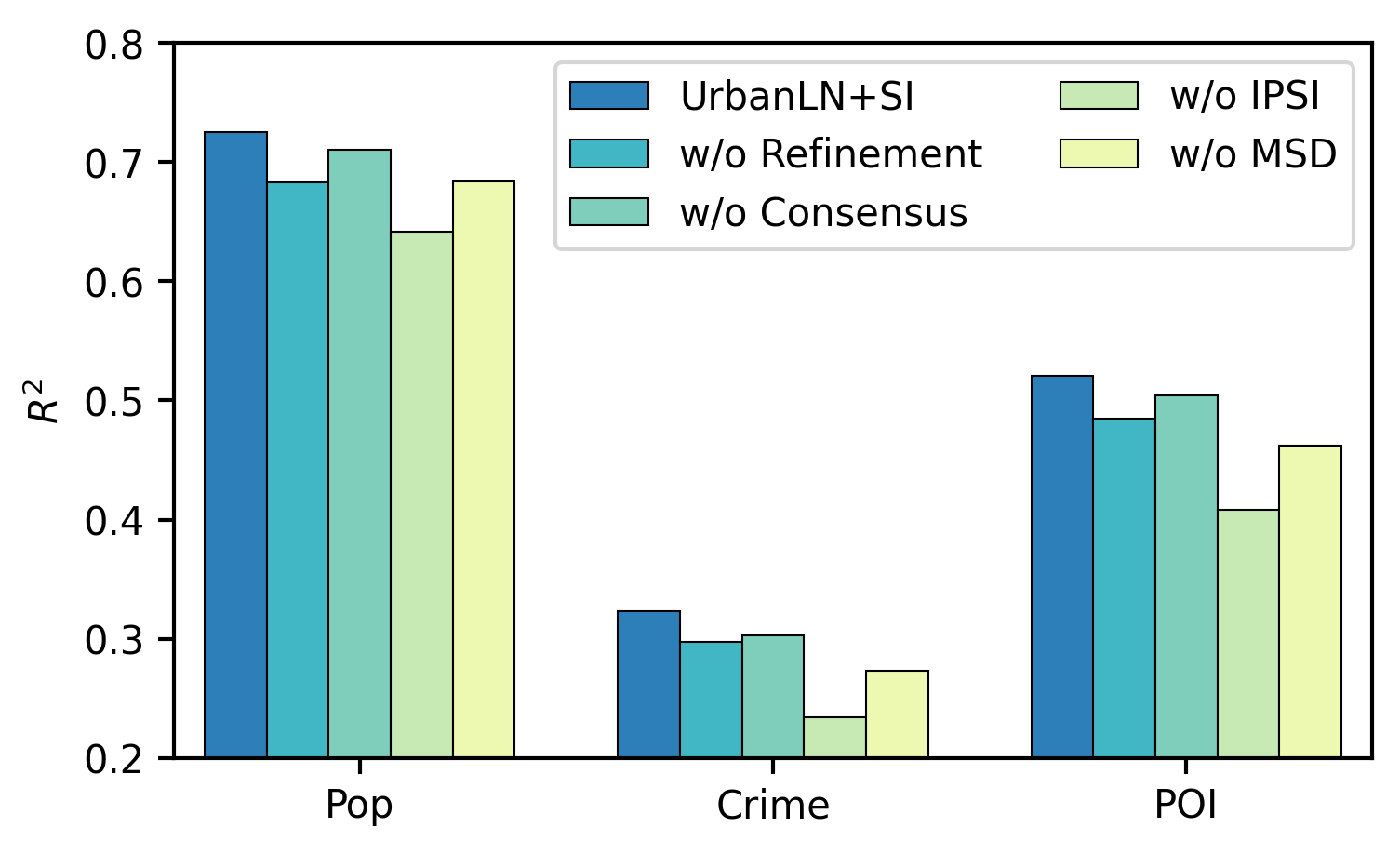}
		\caption{Satellite imagery on NY.}
	\end{subfigure}
	\caption{Results of ablation study on $R^2$ metric.}
	\label{fig_ab}
\end{figure}

\subsection{Model Analysis}

\subsubsection{Ablation Study.} To evaluate the effectiveness of the key components in UrbanLN, we conduct ablation studies across four critical variants. Results are presented in Figure~\ref{fig_ab}, and summarized as follows: \textbf{1) w/o Refinement}: Removing the divide-and-conquer refinement module results in a 10.45\% average performance drop, which highlights the importance of visual verification, local detail enhancement, and hallucination filtering in improving caption quality. \textbf{2) w/o Consensus}: Replacing the consensus-based caption selection with random choice leads to inferior performance, supporting our hypothesis that inter-model agreement serves as a reliable proxy for caption quality in the absence of ground-truth references. \textbf{3) w/o IPSI}: Omitting the IPSI strategy causes the most significant performance degradation across all scenarios, with an average reduction of 26.45\%, marking its critical role in handling long-text inputs. By preserving rich information during knowledge transformation, this method ensures high semantic fidelity, which is particularly essential for complex urban scenarios. \textbf{4) w/o MSD}: Removing the momentum-based self-distillation module results in a notable performance drop, underscoring the importance of generating pseudo-targets in guiding the student model to learn semantically invariant representations under noisy conditions.

\subsubsection{Hyperparameter Analysis.} We explore the impact of pre-training loss weight $\mu$ on model performance, with results presented in Table \ref{tab:mu_impact_comparison}. UrbanLN shows minimal performance variation across different values of $\mu$, but achieves a slight improvement when $\mu=0.5$. Therefore, we adopt $\mu=0.5$ as the default setting.

\begin{table}[t]
	\centering
	\small
	\begin{tabular}{cccccc}
		\toprule
		\multirow{2}{*}{Imagery} & \multirow{2}{*}{$\mu$} & \multicolumn{4}{c}{City} \\
		\cmidrule(lr){3-6}
		& &BJ&SH & SZ & NY \\
		\midrule
		\multirow{4}{*}{Street-view} 
		& 0.3 & 0.687 & 0.655 & 0.662&0.658\\
		& 0.4 & 0.697 & 0.666 & 0.678 & 0.669 \\
		& 0.5 & \textbf{0.705} & \textbf{0.669} & \textbf{0.681} & \textbf{0.676} \\
		& 0.6 & 0.692 & 0.655 & 0.678 & 0.674 \\
		%		\addlinespace[0.5em]
		\midrule
		\multirow{4}{*}{Satellite} 
		& 0.3 & 0.663 & 0.654 & 0.659 & 0.702 \\
		& 0.4 & 0.679 & 0.660 & 0.673 & 0.710 \\
		& 0.5 & \textbf{0.686} & 0.661 & 0.677 & \textbf{0.725} \\
		& 0.6 & 0.682 &\textbf{ 0.665} & \textbf{0.678} & 0.721 \\
		\bottomrule
	\end{tabular}
	\caption{Impact of hyperparameter $\mu$ on population prediction across four cities on $R^2$ metric.}
	\label{tab:mu_impact_comparison}
\end{table}

\subsubsection{Model Scale and Inference Efficiency.} 
\begin{figure}[t]
	\centering
	\includegraphics[width=\linewidth]{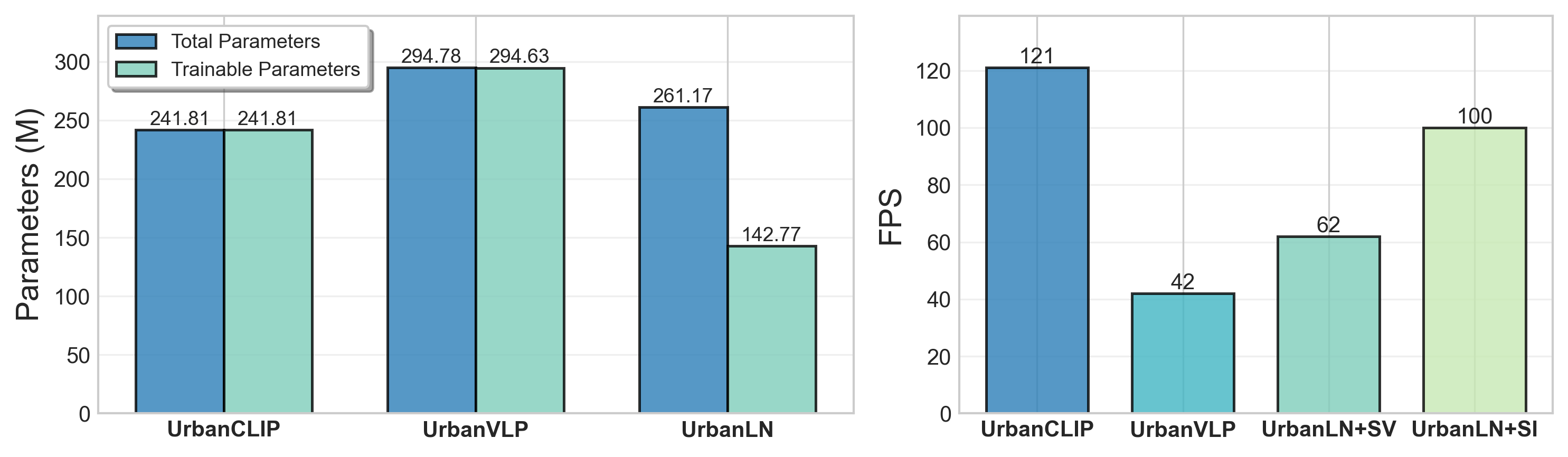}
	\caption{Comparison of parameters and inference speed.}
	\label{Time_Meme}
\end{figure}

We compare the parameter scale and inference efficiency of UrbanLN with two strong baselines: UrbanCLIP and UrbanVLP. Since the caption generation pipeline is a part of data preprocessing, it is excluded from the analysis. As illustrated in Figure \ref{Time_Meme}, all three models maintain comparable total parameter counts. However, UrbanLN substantially reduces the number of trainable parameters. This advantage can be attributed to the strategic integration of the pretrained CLIP model, which allows effective adaptation to urban region representations with minimal fine-tuning. For inference speed, we adopt Frames Per Second (FPS) as the evaluation metric, which is widely used in the computer vision area. While UrbanCLIP achieves the highest FPS, it does so at the expense of prediction accuracy. In contrast, UrbanLN strikes a more favorable balance between accuracy and efficiency, underscoring its potential for real-time urban applications.

\subsection{Transferability Testing}
\begin{figure}[t]
	\centering
	\begin{subfigure}[t]{0.22\textwidth}  
		\includegraphics[width=\linewidth]{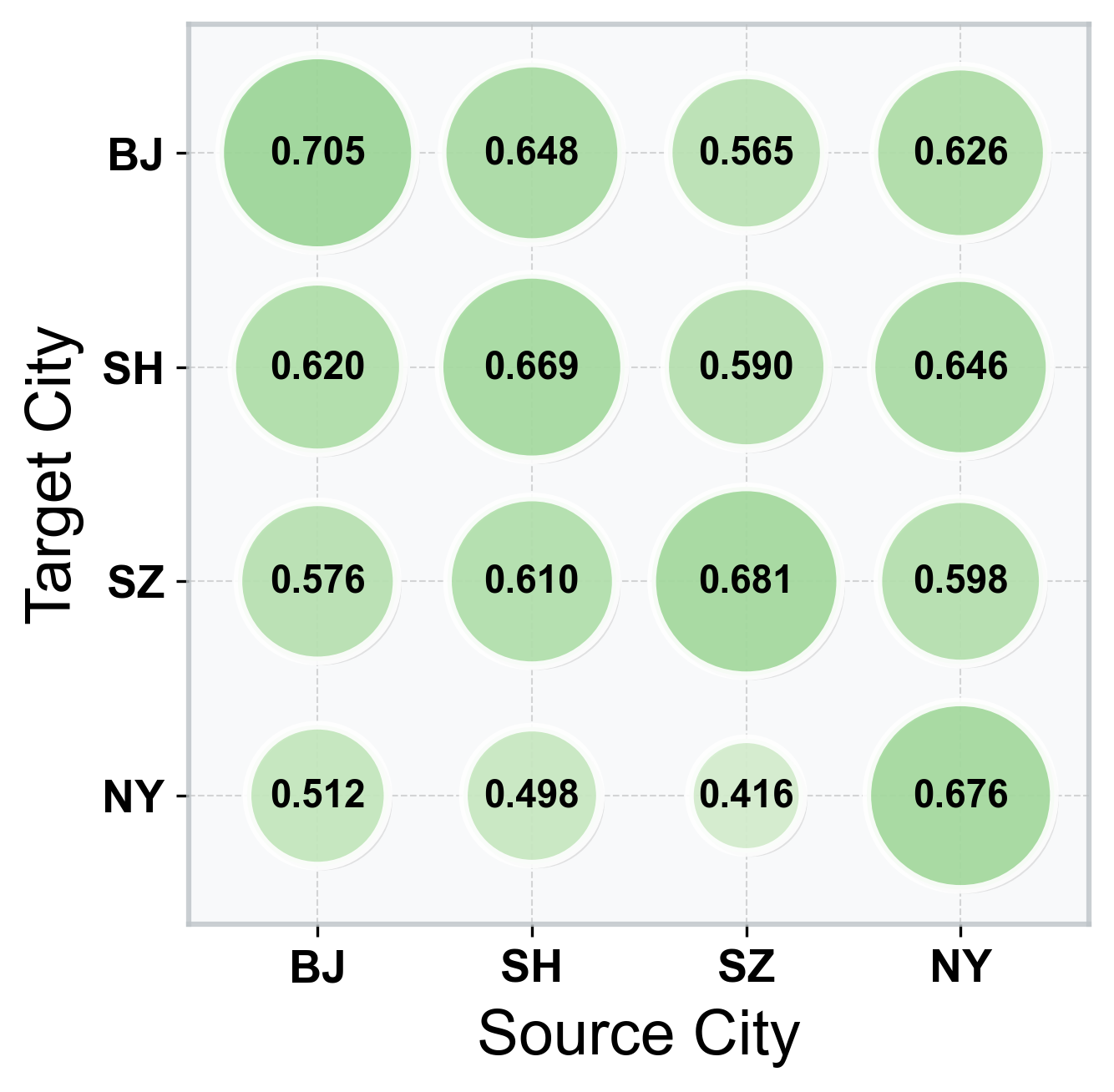}
		\caption{Street-view Imagery}
	\end{subfigure}
	\hfill  
	\begin{subfigure}[t]{0.22\textwidth}
		\includegraphics[width=\linewidth]{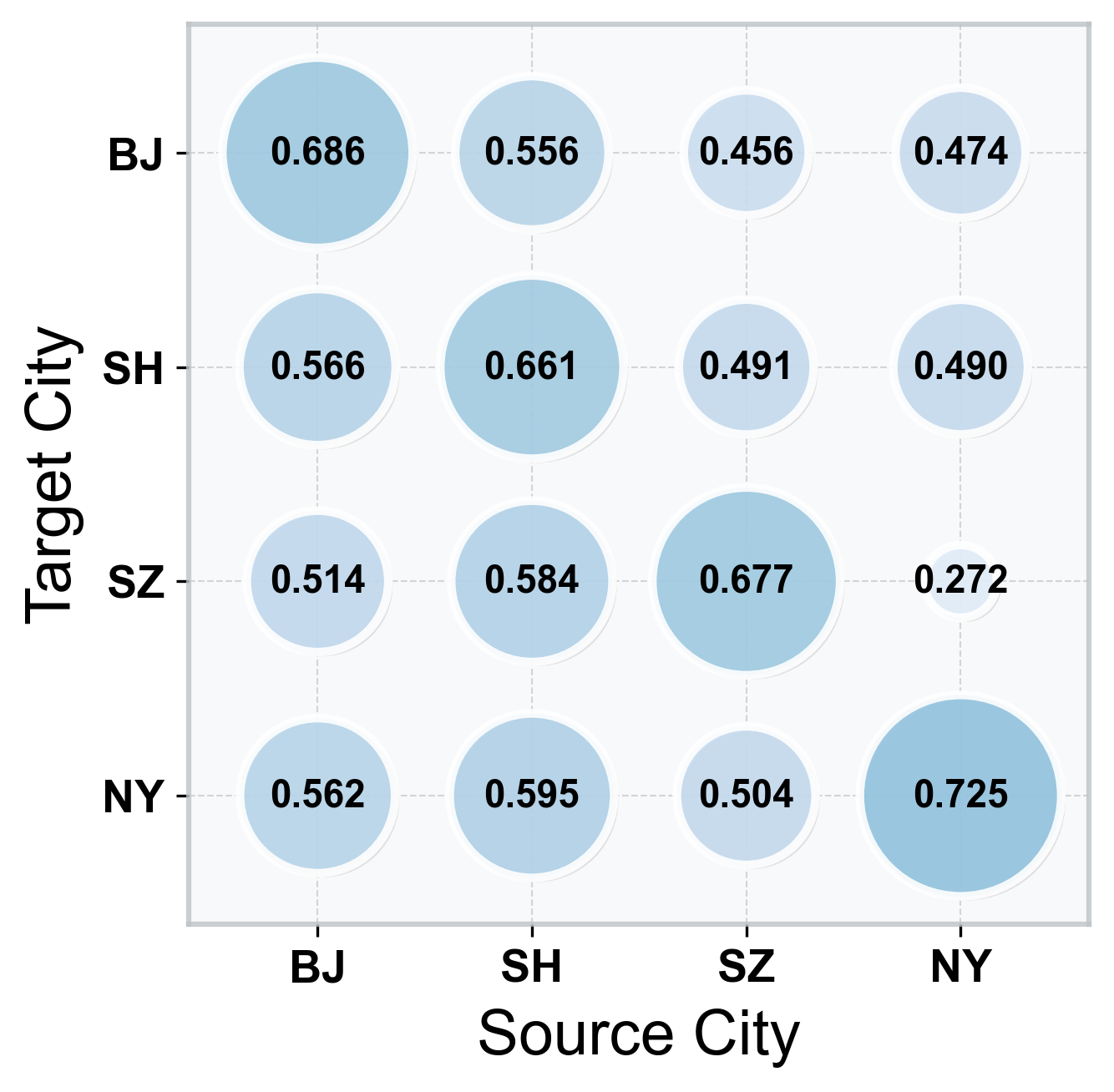}
		\caption{Satellite Imagery}
	\end{subfigure}
	\caption{The $R^2$ for the transferability test on street-view and satellite imagery-based population prediction.}
	\label{fig5}
\end{figure}

To assess the capability of our model in the practical case, we pre-train the UrbanLN in a source city and evaluate its transferability by applying it to other cities. The experimental results presented in Figure~\ref{fig5} indicate that, despite the inherent differences among cities, the model seems to capture fundamental commonalities across diverse urban environments, consistently achieving high prediction accuracy. This performance can be attributed to the effective alignment of visual and textual representations, allowing the model to extract universal functional semantics from urban images.

\subsection{Latent Space Exploration}
This case study aims to disentangle the information encoded in urban region representations and provide an intuitive illustration of the model’s capabilities in modeling city features. Using Shenzhen as an example, we first project the learned representations via principal component analysis and apply K-means clustering to partition the city into three distinct categories. Then, we perform a density-based analysis using population data to examine the characteristic patterns of each cluster. 

\begin{figure}[t]
	\centering
	\includegraphics[width=\linewidth]{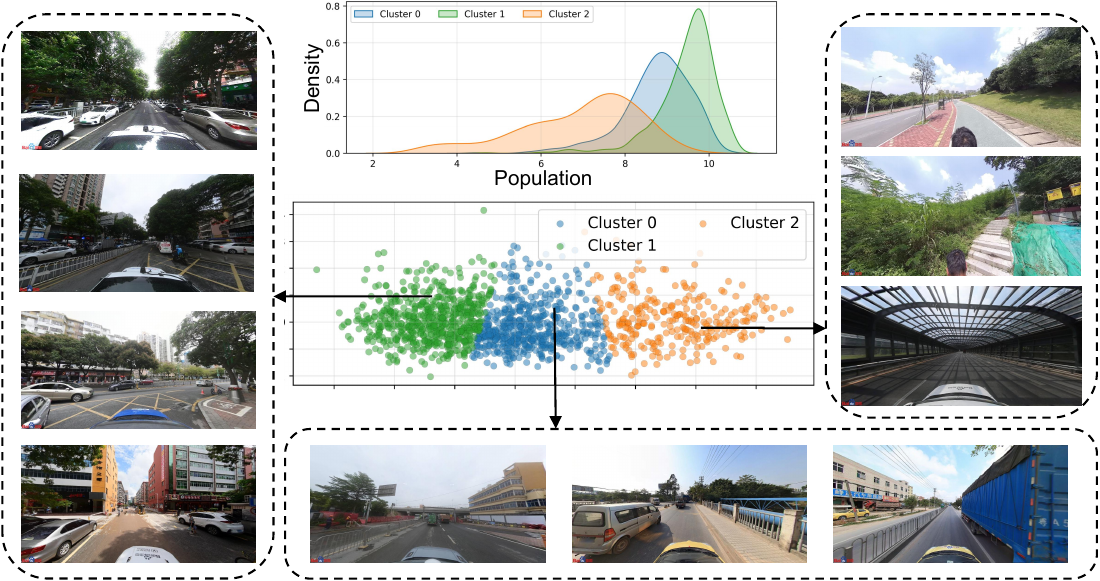}
	\caption{Case study on SZ using street-view imagery.}
	\label{fig9}
\end{figure}

As shown in Figure~\ref{fig9}, the clusters exhibit clear spatial distinctions aligned with urban development patterns. Cluster 1 (green) shows a sharp, narrow peak in the KDE curves, indicating dense central urban areas with high population. Street-view images from this cluster reveal high-rise buildings, busy streets, and commercial infrastructure, supporting this interpretation. Cluster 0 (blue), which overlaps with Clusters 1 and 2, likely represents transitional zones between urban cores and outlying areas. Cluster 2 (orange) corresponds to suburban or rural regions with lower population. These findings demonstrate the model’s ability to capture meaningful socio-economic structures within urban environments.

\section{Related Work}
Early studies primarily relied on supervised learning for specific tasks \cite{yeh2020using}. However, these ``one-to-one'' approaches not only require large amounts of labeled data but also suffer from limited transferability. To address this limitation, recent studies \cite{li2024urban,zhang2022region, sun2025flexireg} have shifted toward self-supervised representation learning for urban region analysis. This general-purpose approach eliminates the need to train specialized models for each downstream task, thereby enhancing efficiency and improving model generalizability \cite{liang2025foundation}.

Satellite imagery serves as a crucial visual data source for studying the spatial characteristics of urban regions and has been widely utilized in urban understanding \cite{burke2021using}. PG-SimCLR \cite{xi2022beyond} and ReFound \cite{xiao2024refound} leveraged a combination of visual information from satellite imagery and semantic information from POIs to construct urban region representations. UrbanCLIP \cite{yan2024urbanclip} introduced textual descriptions as a complementary modality to enhance the visual information contained in images. Moreover, street-view images have gained popularity in urban understanding due to their rich visual details. Vision-LSTM \cite{huang2023comprehensive} used an LSTM-based network to integrate diverse street images into the satellite image domain, while UrbanVLP \cite{hao2025urbanvlp} integrated two visual modalities through token-level contrastive learning to capture multi-granularity information. 

\section{Conclusion}
In this work, we propose UrbanLN, a cross-modal pre-training framework designed to improve urban region representation by addressing two key challenges in LLM-guided urban imagery analysis: the difficulty of extracting fine-grained semantics from long-form textual descriptions and the presence of noise in LLM-generated captions. To this end, UrbanLN introduces a long-text-aware strategy to preserve detailed semantic alignment and a hierarchical noise mitigation mechanism operating at both the data and model levels. The experiment results demonstrate that our model achieves state-of-the-art performance across various downstream tasks in four real-world datasets.

\section{Ethical Statement}
This study uses satellite imagery and street-view images obtained from the Baidu platform solely for academic research purposes. The spatial resolution of the satellite imagery is insufficient to identify individuals and is used solely to detect environmental changes associated with human activities.  Street-view images were crawled and downloaded from publicly accessible content on the platform, with key identifying elements blurred to ensure that no private information is compromised. The research did neither involve interventions with human subjects nor individualized human data. Thus, no approval from the Institutional Review Board was required by the author institutions.

\section{Acknowledgments}
This work was supported in part by the National Natural Science Foundation of China under Grant 62476247, 62073295 and 62072409,  ``Pioneer" and ``Leading Goose" R\&D Program of Zhejiang under Grant 2024C01214, in part by the Zhejiang Provincial Natural Science Foundation under Grant LR21F020003, in part by the Supcon Research Fund under Grant KYY-HX-20230833, and Lantai Research Fund under Grant KYY-HX-20240573, KYY-HX-20230365, KYY-HX-20250588.

\bibliography{aaai2026}

\clearpage
\appendix
\section{A Appendix}

\subsection{A.1  More Details about Datasets}
Following prior works such as UrbanCLIP and UrbanVLP, we adopt urban imagery from Beijing, Shanghai, and Shenzhen as part of our experimental datasets. To further assess the generalization capability of our model across diverse urban environments, we replace the original city of Guangzhou with New York City. This choice introduces greater geographical and cultural variance.
\subsubsection{Urban Imagery.}  We collect street-view and satellite images for Beijing, Shanghai, and Shenzhen using Baidu Maps\footnote{lbsyun.baidu.com/}. For street-view imagery, we first obtain the road network data from OpenStreetMap\footnote{openstreetmap.org/}, and generate sampling points at fixed 500-meter intervals along the road network. For each sampling point, we retrieve the corresponding street-view image from Baidu Maps using its latitude and longitude coordinates, with a resolution of 1024×512 pixels. Due to occasional unavailability of images at certain locations, the final number of street-view images is smaller than the total number of sampling points. Satellite imagery consists of 3-channel RGB images at a resolution of 256×256 pixels. For New York City, both street-view and satellite images are obtained from the dataset released in \cite{yong2024musecl}.
\subsubsection{Downstream Tasks.} We evaluate our model on five downstream tasks in Beijing and Shenzhen, and four in Shanghai (excluding the comment dataset due to data unavailability). The tasks include:
\begin{itemize}
	\item Population (\textbf{Pop}): Derived from the 2020 WorldPop dataset\footnote{worldpop.org/}, with units measured as number of citizens.
	\item Gross Domestic Product (\textbf{GDP}): Collected from \cite{zhao2017forecasting} to represent regional economic development, measured in million Chinese Yuan.
	\item Nighttime Light Intensity (\textbf{Night}): Sourced from \cite{zhong2022development}, capturing urban activity levels and development patterns.
	\item Restaurant Comments (\textbf{Com}): Obtained from Dianping platform via \cite{dong2019predicting}, reflecting residents' consumption activity.
	\item Carbon Emissions ($\mathbf{CO_2}$): Extracted from the ODIAC database\footnote{db.cger.nies.go.jp/dataset/ODIAC/}, used as an environmental indicator. Emissions are reported in annual tons.
\end{itemize}
For New York City, we follow the setting of \cite{yong2024musecl}. Regions with missing data were excluded to ensure dataset integrity. Summary statistics of the imagery and downstream indicators are provided in Table~\ref{table_data}.

\begin{table*}[t]
	\centering
	\renewcommand{\arraystretch}{1.2}
	\begin{tabular}{@{}l*{10}{c}@{}}
		\toprule
		\multirow{2}{*}{\textbf{City}} & \multirow{2}{*}{\textbf{\# Region}} &\multicolumn{2}{c}{\textbf{Urban Imagery}} & \multicolumn{7}{c}{\textbf{Downstream Tasks}} \\
		\cmidrule(lr){3-4} \cmidrule(lr){5-11}
		& & \textbf{\# SV} & \textbf{\# SI} & \textbf{\# Pop} & \textbf{\# GDP} & \textbf{\# Night} & \textbf{\# Com} & \textbf{\# CO$_2$} & \textbf{\# Crime} & \textbf{\# POI} \\
		\midrule
		\textbf{BJ}  &4,584 & 29,814 & 4,584  & 3,168  & 2,056 & 3,181 & 2,188 & 2,704 & --   & --   \\
		\textbf{SH}  &5,168 & 37,469 & 5,168 & 3,726  & 2,876 & 3,730 & --   & 3,513 & --   & --   \\
		\textbf{SZ} &3,383  & 33,897 & 3,383 & 1,450  & 1,274 & 1,484 & 1,052 & 1,444 & --   & --   \\
		\textbf{NY} &517    & 44,617 & 517   & 506  & --   & --   & --   & --   & 422   & 516   \\
		\bottomrule
	\end{tabular}
	\caption{Statistical Information of Datasets. ``--" indicates unavailable data. In the downstream task, we count the number of final valid test regions.}
	\label{table_data}
\end{table*}

\subsection{A.2  Details of Baselines}
We compare our method with the following baseline methods:
\begin{itemize}
	\item ResNet-18 \cite{he2016deep}: It is a well-established deep learning model pre-trained on ImageNet to generate visual representations of images.
	\item Tile2Vec \cite{jean2019tile2vec}: Tile2Vec is an unsupervised representation learning model designed for geospatial imagery. It is based on the assumption that geographically adjacent image tiles tend to exhibit similar semantics, while distant tiles are likely to possess dissimilar semantics. To capture this spatial semantic relationship, Tile2Vec employs a triplet loss that minimizes the distance between representations of neighboring tiles while maximizing the distance between distant ones.
	\item PG-SimCLR \cite{xi2022beyond}: PG-SimCLR extends the traditional SimCLR framework by incorporating both satellite imagery and POI data to capture geographic proximity and human activity factors in the learned representations. It introduces a contrastive learning framework that fuses POI information with satellite imagery, thereby enriching the learned representations of urban regions.  
	\item RemoteCLIP \cite{liu2024remoteclip}: RemoteCLIP is the first vision-language foundation model for remote sensing that aims to learn robust visual features with rich semantics and aligned text embeddings for seamless downstream application.
	\item UrbanCLIP \cite{yan2024urbanclip}: UrbanCLIP is the first LLM-enhanced framework that integrates textual knowledge into urban imagery analysis. It adopts an image-text contrastive learning approach to incorporate domain-specific knowledge from text modalities, thereby enhancing the model’s ability to extract robust and semantically rich representations from urban imagery. 
	\item MuseCL \cite{yong2024musecl}: MuseCL achieves fine-grained urban region profiling by jointly exploring visual and textual semantics. It uses both street-view and satellite imagery, concurrently integrating POI and mobility flow data to enrich the embedding with multi-dimensional semantic information. Due to the lack of mobility data in BJ, SH, and SZ, the comparative analysis with the MuseCL baseline was conducted solely on the NY dataset. 
	\item UrbanVLP \cite{hao2025urbanvlp}: UrbanVLP explores the semantic granularity differences between street-view and satellite images and investigates their complementary roles in modeling urban region representations. It introduces a text generation and calibration mechanism to ensure the generation of high-quality textual descriptions.
\end{itemize}

\subsection{A.3 Additional Experimental Results}
The experimental results of UrbanLN across the SH, SZ, and NY datasets are comprehensively summarized in Tables \ref{table_SH}, \ref{table_SZ}, and \ref{table_NY}. UrbanLN consistently demonstrates superior performance across all downstream tasks on the SH dataset, showcasing exceptional performance. On the SZ dataset, UrbanLN achieves state-of-the-art results in 14 out of 15 entries. On the NY dataset, UrbanLN+SV consistently outperforms all competing methods, achieving substantial performance improvements ranging from 4.3\% to 54.8\%. UrbanLN+SI exhibits strengths in population-related tasks. However, its performance is comparatively weaker in crime prediction and POI prediction tasks. This limitation is attributed to inherent bottlenecks in satellite image data, as elaborated in the main text. Despite these challenges, UrbanLN+SI still surpasses competing satellite-only baselines (e.g., RemoteCLIP and UrbanCLIP).

\begin{table*}[t]
	\footnotesize
	\centering
	\begin{tabular}{l|ccc|ccc|ccc|ccc}
		\toprule 
		%		\rowcolor{gray!20}
		\multirow{1}{*}{\textbf{Task $\rightarrow$}}  & \multicolumn{3}{c}{\textbf{Pop}}   & \multicolumn{3}{c}{\textbf{GDP}} & \multicolumn{3}{c}{\textbf{Night}} & \multicolumn{3}{c}{\textbf{CO$_2$}} \\ 	\cmidrule(lr){2-4} \cmidrule(lr){5-7} \cmidrule(lr){8-10} \cmidrule(lr){11-13} 
		%		\rowcolor{gray!20}
		\multirow{1}{*}{\textbf{Model $\downarrow$}}           & $R^2$ & RMSE& MAE & $R^2$ & RMSE& MAE & $R^2$ & RMSE & MAE & $R^2$ & RMSE & MAE            \\ \midrule
		ResNet-18	&0.454 &	0.940 	&0.779 	 &	0.281 &	1.462 &	1.170 &	0.275 &	0.835 	&0.602 &	0.454 &	0.650 &	0.478 \\
		Tile2Vec&	0.482 &	0.836 &	0.632 &	0.351 &	1.396 &	1.161 &	0.316 &	0.609 &	0.457 	&0.542 &	0.585 &	0.442 \\ 
		PG-SimCLR&	0.544 &	0.827 &	0.634  &	0.472 &	1.266 &	0.995 &	0.348 &	0.575 &	0.459 	&0.568 &	0.552 &	0.434 \\
		RemoteCLIP&	0.524 &	0.777 &	0.593 	&	0.395 &	1.397 &	0.999 &	0.391 &	0.747 &	0.582&	0.564 &	0.644 &	0.455 \\
		UrbanCLIP&	0.554 &	0.707 &	0.542  &	\textit{0.473} &\underline{1.213} &\textit{0.994} &	0.464 &	0.500 &	0.395 &0.615 &	0.521 &	0.410 \\
		UrbanVLP &  \textit{0.576} &\textit{0.689} &	\textit{0.532}  &	0.458 &	1.259 &	0.995 &	\textit{0.506} &\textit{0.480} &\textit{0.378}  &\textit{0.637} &\textit{0.506} &	\textit{0.396}            \\ 
		
		\midrule
		\textbf{UrbanLN+SV}&  \textbf{0.669} &\textbf{0.599} &	\textbf{0.470} 		&\textbf{0.480} 	&\underline{1.213} 	&\textbf{0.981} 	&\textbf{0.552} 	&\textbf{0.458} 	&\textbf{0.360} &\underline{0.697} 	&\underline{0.464} 	&\underline{0.362} \\

		Improvement &  16.1\%	&13.1\%	&11.7\%		&1.5\%	&0.0\%	&1.3\%	&9.1\%	&4.6\%	&4.8\%	&9.4\%	&8.3\%	&8.6\% \\
		
		\textbf{UrbanLN+SI}   &\underline{0.661} 	&\underline{0.607} 	&\underline{0.473} 		&\underline{0.476} 	&\textbf{1.212} 	&\underline{0.985} 	&\underline{0.555} 	&\underline{0.460} 	&\underline{0.364} &\textbf{0.718} 	&\textbf{0.458} 	&\textbf{0.356} 
		\\
		
		Improvement &  14.8\%	&11.9\%	&11.1\%	&0.6\%	&3.7\%	&1.0\%	&9.7\%	&4.2\%	&3.7\%	 &12.7\%	&9.5\%	&10.1\%  
		\\ 
		\bottomrule
	\end{tabular}
	\caption{Detailed prediction results of downstream tasks on SH the dataset. }
	\label{table_SH}
\end{table*}

\begin{table*}[t]
	\footnotesize
	\setlength{\tabcolsep}{3pt}
	\centering
	\begin{tabular}{l|ccc|ccc|ccc|ccc|ccc}
		\toprule 
		%		\rowcolor{gray!20}
		\multirow{1}{*}{\textbf{Task $\rightarrow$}}  & \multicolumn{3}{c}{\textbf{Pop}}  & \multicolumn{3}{c}{\textbf{GDP}} & \multicolumn{3}{c}{\textbf{Night}} & \multicolumn{3}{c}{\textbf{Com}}  & \multicolumn{3}{c}{\textbf{CO$_2$}} \\ 	\cmidrule(lr){2-4} \cmidrule(lr){5-7} \cmidrule(lr){8-10} \cmidrule(lr){11-13} \cmidrule(lr){14-16} 
		%		\rowcolor{gray!20}
		\multirow{1}{*}{\textbf{Model $\downarrow$}}           & $R^2$ & RMSE& MAE & $R^2$ & RMSE& MAE & $R^2$ & RMSE & MAE & $R^2$ & RMSE & MAE & $R^2$ & RMSE & MAE        \\ \midrule
		ResNet-18	&0.097 	&1.094 	&0.852 	&0.154 	&1.916 	&1.542 	&0.277 	&0.580 	&0.453 	&0.133 	&2.645 	&2.192 	&0.201 	&0.802 	&0.572 \\
		Tile2Vec	&0.245 	&1.032 	&0.792 	&0.184 	&1.763 	&1.441 	&0.280 	&0.575 	&0.446 	&0.296 	&2.409 	&2.019 	&0.397 	&0.656 	&0.499 \\
		PG-SimCLR	&0.285 	&1.035 	&0.786 	&0.245 	&1.770 	&1.464 	&0.426 	&0.529 	&0.418 	&0.359 	&2.268 	&1.821 	&0.574 	&0.543 	&0.411 \\
		RemoteCLIP	&0.262 	&1.052 	&0.775 &	0.165 	&1.885 	&1.537 	&0.365 	&0.557 	&0.425 	&0.232 	&2.481 	&2.046 	&0.371 &	0.660 	&0.502 \\
		UrbanCLIP	&0.524 	&0.929 	&0.668 		&0.300 	&1.712 	&1.407 	&0.490 	&0.498 	&\textit{0.382} 	&0.524 	&1.897 	&1.473 	&0.576 	&\textit{0.542} 	&\textbf{0.392} \\
		
		UrbanVLP &  \textit{0.542} 	& \textit{0.828} 	&\textit{0.617} 	& \textit{0.306} 	&\textit{1.605} 	&\textit{1.296} 	& \textit{0.513} 	&\textit{0.493} 	& 0.386 	&\textit{0.556} 	&\textit{1.833} 	& \textit{1.429} 	& \textit{0.649} 	& 0.546 	& 0.416  \\ 
		
		\midrule
		\textbf{UrbanLN+SV}& \textbf{0.681} 	&\textbf{0.765} 	&\textbf{0.552} 		&\textbf{0.311} 	&\textbf{1.596} 	&\textbf{1.287} 	&\textbf{0.615} 	&\textbf{0.437} 	&\underline{0.325} 	&\textbf{0.639} 	&\textbf{1.718} 	&\underline{1.314} 	&\underline{0.679} 	&\underline{0.523} 	&\textit{0.402} \\
		
		Improvement &  25.6\%	&7.6\%	&10.5\%		&1.6\%	 &0.6\%	&0.7\%	 &19.9\%	&12.2\%	&14.9\%	&14.9\%	&6.3\%	&8.0\%	 &4.6\%	&3.5\%	 &-2.6\% \\
		
		\textbf{UrbanLN+SI}   &\underline{0.677} 	&\underline{0.772} 	&\underline{0.554} 	&\underline{0.307} 	&\underline{1.602} 	&\underline{1.291} 	&\underline{0.601} 	&\underline{0.443} 	&\textbf{0.319} 	&\underline{0.622} 	&\underline{1.724} 	&\textbf{1.311} 	&\textbf{0.715} 	&\textbf{0.519} 	&\underline{0.396} 	\\
		
		Improvement &  24.9\%	&6.8\%	&10.2\%	&0.3\% 	&0.2\%	 &0.4\%	&17.2\%	&11.0\%	&16.5\%	&11.9\%	&5.9\%	&8.3\%	&10.2\%	&4.2\%	&-1.0\%
		\\ 
		\bottomrule
	\end{tabular}
	\caption{Detailed prediction results of downstream tasks on the SZ dataset. }
	\label{table_SZ}
\end{table*}

\begin{table*}[t]
	\footnotesize
	\setlength{\tabcolsep}{6pt}
	\centering
	\begin{tabular}{l|ccc|ccc|ccc}
		\toprule 
		%		\rowcolor{gray!20}
		\multirow{1}{*}{\textbf{Task $\rightarrow$}}  & \multicolumn{3}{c}{\textbf{Pop}}   & \multicolumn{3}{c}{\textbf{Crime}} & \multicolumn{3}{c}{\textbf{POI}}  \\ 	\cmidrule(lr){2-4} \cmidrule(lr){5-7} \cmidrule(lr){8-10} 
		%		\rowcolor{gray!20}
		\multirow{1}{*}{\textbf{Model $\downarrow$}}           & $R^2$ & RMSE& MAE & $R^2$ & RMSE& MAE & $R^2$ & RMSE & MAE          \\ \midrule
		ResNet-18	&-0.079 	&2.154 	&1.528 		&0.050 	&1.520 	&1.280 	&0.160 	&1.217 	&0.870 \\
		Tile2Vec	&0.096 	&2.073 	&1.510 	&0.035 &	1.586 &	1.297 	&0.192 	&1.154 	&0.832 \\
		PG-SimCLR	&0.375 	&1.942 	&1.324 		&0.196 	&1.337 	&1.158 		&-- &-- &--	\\
		RemoteCLIP	&0.364 &	1.995 	&1.397 	&0.154 	&1.453 	&1.225 	&0.328 	&1.096 	&0.824 \\
		UrbanCLIP	&0.448 	&1.827 	&1.268 	&0.251 	&1.324 	&1.173 	&0.425 	&0.973 	&0.785 \\
		MuseCL &0.521 	&1.789 	&1.053 	&\textit{0.368} 	&\textit{1.196} 	&\textit{0.990} 	&\textit{0.542} 	&\underline{0.881} 	&\underline{0.701} \\
		UrbanVLP &  \textit{0.534} 	&\textit{1.623} 	&\textit{0.926} 	 	&\underline{0.467} 	&\underline{1.147}	&\underline{0.916}	&\underline{0.583} 	&\textit{0.932} 	&\textit{0.759} 	  \\ 
		\midrule
		\textbf{UrbanLN+SV} & \underline{0.676}	&\underline{0.994}	&\underline{0.691}		&\textbf{0.723}	&\textbf{0.841} 	&\textbf{0.665} 	&\textbf{0.650}	&\textbf{0.824} 	&\textbf{0.671} 	\\
		Improvement &  26.6\%	&38.8\%	 &25.4\%  &54.8\%	&26.7\%	 &27.4\%	&11.5\%	 &6.5\%	 &4.3\%\\
		
		\textbf{UrbanLN+SI}   &\textbf{0.725} 	&\textbf{0.912} 	&\textbf{0.680} 	&0.323 	&1.315 	&1.126 	&0.521 	&0.965 	&0.767 	\\
		
		Improvement &  35.8\%	& 43.8\%	& 26.6\%	& -30.8\%	  & -14.6\%	& -22.9\%	& -10.6\%	& -9.5\%	& -9.4\%
		\\ 
		\bottomrule
	\end{tabular}
	\caption{Detailed prediction results of downstream tasks on the NY dataset. }
	\label{table_NY}
\end{table*}

\subsection{A.4 UrbanLN vs. UrbanCLIP and UrbanVLP}
\begin{figure}[t]
	\centering
	\includegraphics[width=\linewidth]{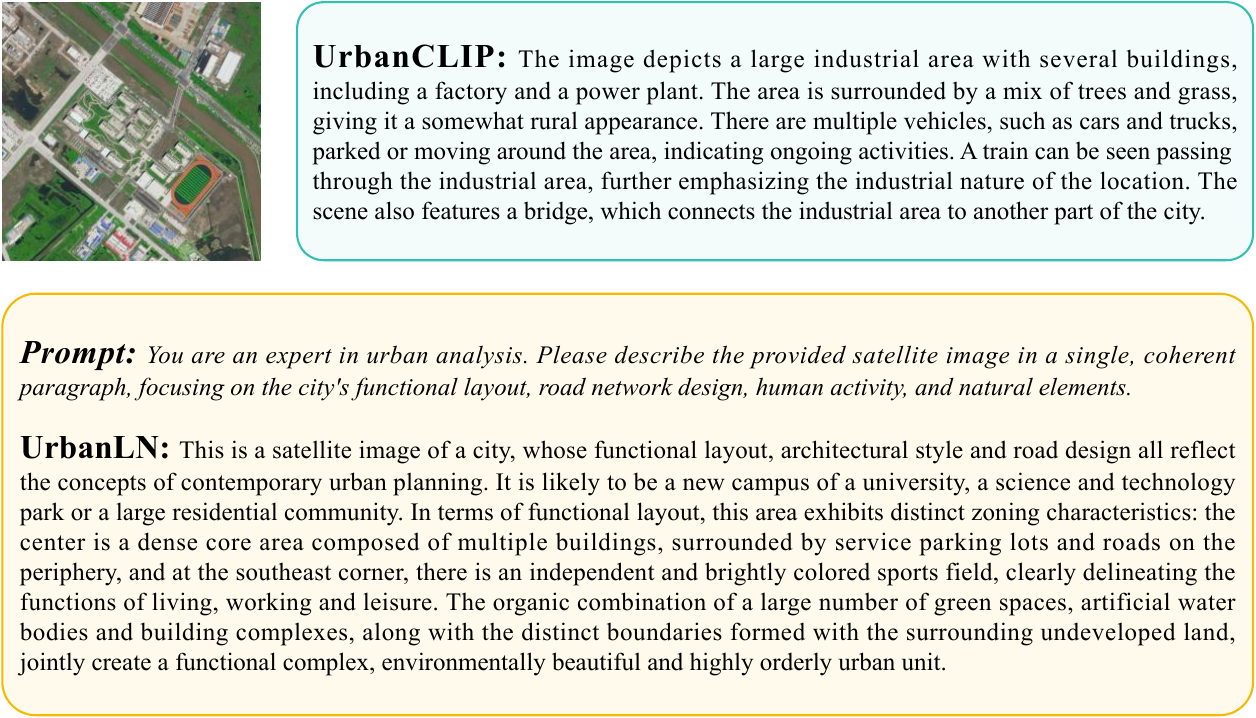}
	\caption{Comparison of satellite image captions generated by UrbanCLIP and UrbanLN.}
	\label{satellite_caption}
\end{figure}

\begin{figure}[t]
	\centering
	\includegraphics[width=\linewidth]{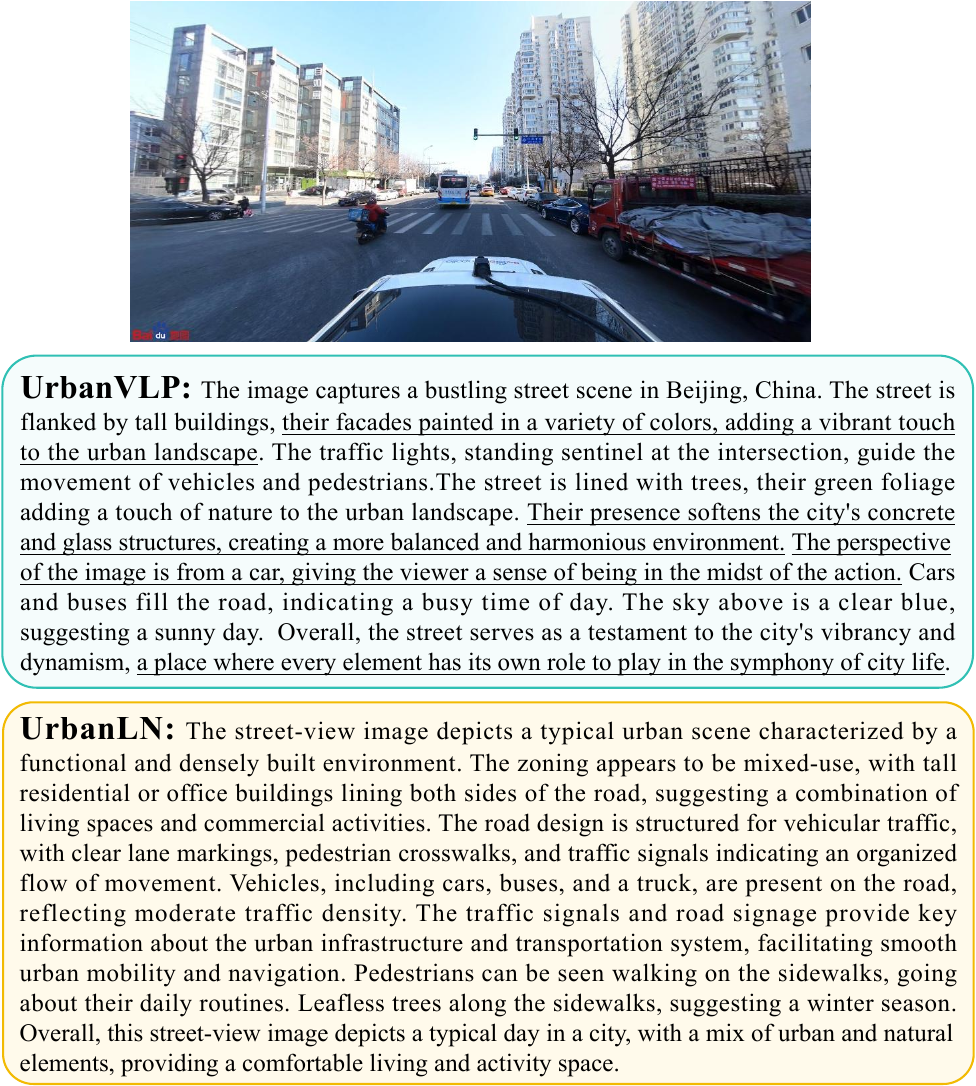}
	\caption{Comparison of street-view image captions generated by UrbanVLP and UrbanLN.}
	\label{street_caption}
\end{figure}

In this section, we present a more detailed analysis of two studies that are most closely related to our work. At the data level, UrbanLN supports automatic caption generation, enabling scalable across large datasets. Furthermore, it facilitates the generation of diverse and semantically rich textual descriptions. Below, we provide two specific examples to qualitatively compare the generated captions of our UrbanLN with UrbanCLIP and UrbanVLP. Figure~\ref{satellite_caption} presents the captions generated by UrbanCLIP and UrbanLN for the same satellite image. It can be seen that compared with manually calibration in UrbanCLIP, our automatic generation and refinement method can also achieve comparable performance. Figure~\ref{street_caption} presents a comparison of captions generated by UrbanVLP and UrbanLN for the same street-view image. UrbanVLP relies on a single MLLM and adopts a coarse-grained text calibration strategy that integrates a text-to-image model and a segmentation model with 13 predefined categories. Consequently, it often produces generic, stylistically homogeneous descriptions with relatively low information density. In contrast, UrbanLN leverages multiple MLLMs and employs a fine-grained hierarchical refinement process that iteratively operates from global to local and back to global levels. This design explicitly preserves linguistic diversity and enables the model to more effectively mitigate visual omissions, semantic drift, and hallucinations, ultimately generating more accurate, expressive, and context-aware captions. 

\end{document}